\PassOptionsToPackage{table}{xcolor}

\documentclass[10pt,twocolumn,letterpaper]{article}

\usepackage[pagenumbers]{cvpr} %

\newcommand{\ourname}[1]{LaS-Comp}

\usepackage{pifont}
\usepackage{booktabs}
\usepackage{threeparttable} %

\usepackage{tabularx,booktabs,makecell,array,multirow,xcolor}
\newcolumntype{Y}{>{\centering\arraybackslash}X}  %

\usepackage{capt-of}
\usepackage{float}

\usepackage{tabularx,booktabs,makecell,array}
\usepackage[table]{xcolor}      %
\newcolumntype{L}{>{\raggedright\arraybackslash}X}

\newcolumntype{C}{>{\centering\arraybackslash}X}
\usepackage[table]{xcolor}
\usepackage{colortbl} %
\definecolor{bestpink}{HTML}{f9dce0}      %
\definecolor{secondbeige}{HTML}{fff0d8}   %

\usepackage{xcolor}

\usepackage{soul}
\usepackage{color}

\definecolor{cvprblue}{rgb}{0.21,0.49,0.74}
\usepackage[pagebackref,breaklinks,colorlinks,allcolors=cvprblue]{hyperref}

\def\confName{CVPR}
\def\confYear{2027}

\newcommand{\ourattn}{temporal and spatial reference attention}
\newcommand{\ourtta}{latent-reference consistency}
\title{WorldBridge: Bridging Scattered Views for Coherent 3D World Generation}

\author{
Weilong Yan\textsuperscript{1,2,*}\quad
Jingyu Hu\textsuperscript{3,\textdagger}\quad
Hongli Xu\textsuperscript{2,\textdaggerdbl}\quad
Yifan Zhan\textsuperscript{4}\quad
Beibei Lin\textsuperscript{1}\\
Anbang Wang\textsuperscript{2}\quad
Qinqian Lei\textsuperscript{1}\quad
Jingwei Huang\textsuperscript{2}\quad
Robby T.~Tan\textsuperscript{1}\\[0.25em]
{\normalsize
\textsuperscript{1}National University of Singapore\quad
\textsuperscript{2}Tencent}\\
{\normalsize
\textsuperscript{3}The Chinese University of Hong Kong\quad
\textsuperscript{4}The University of Tokyo}
}

\begin{document}
\twocolumn[{
\renewcommand\twocolumn[1][]{#1}
\maketitle
\begin{center}
 \includegraphics[width=.98\linewidth]
{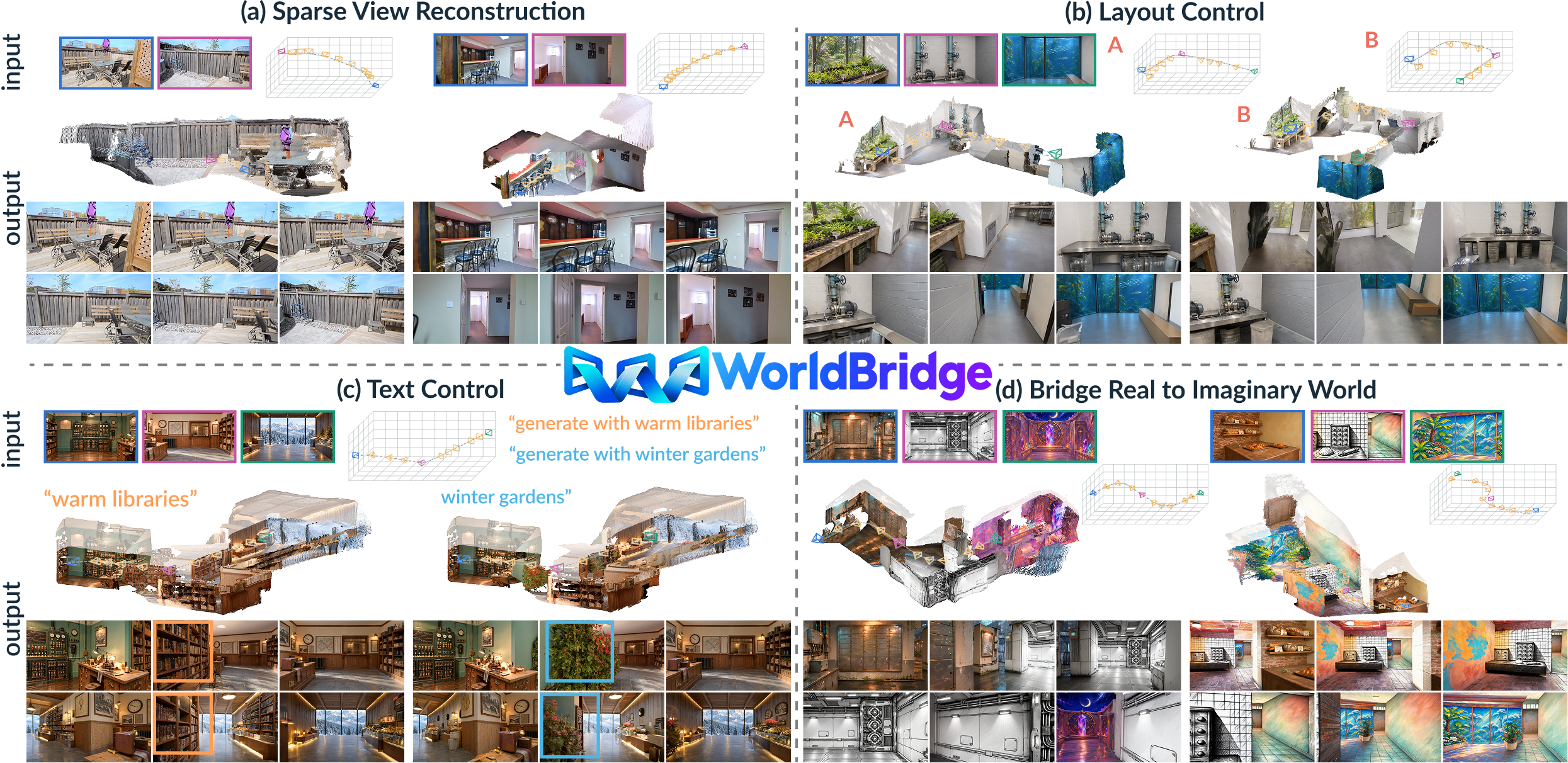}
\makeatletter\def\@captype{figure}\makeatother
\caption{\textbf{WorldBridge} generates coherent 3D worlds from scattered reference images with little or no visual overlap. It preserves observed content while plausibly generating unseen regions, and supports (a) sparse-view scene reconstruction, (b) layout control through reference camera poses, (c) text-guided generation of unseen content, and (d) bridging references spanning real and imaginary visual styles.}
\label{fig:teaser}
\end{center}
\vspace{3pt}
}]
\begingroup
\renewcommand{\thefootnote}{\fnsymbol{footnote}}
\footnotetext[1]{Work done during the internship at Tencent.}
\footnotetext[2]{Corresponding author.}
\footnotetext[3]{Project leader.}
\endgroup

\begin{abstract}

Generating coherent and explorable 3D worlds from scattered images with little or no visual overlap is a fundamental and long-standing challenge in computer vision, which existing geometry-aware methods fail to address.
Meanwhile, existing image- and video-based methods may generate unseen regions but produce inconsistent geometry across views.
To bridge this gap, we present \textbf{WorldBridge}, a framework for coherent and controllable 3D world generation under scattered images with little or no visual overlap.
WorldBridge encodes scattered references into a global scene context to guide geometry and appearance generation across views.
Our \ourattn{} mechanism further enhances the fidelity of temporal references along the target trajectory and supports flexible spatial references outside it.
A plug-and-play latent-reference consistency strategy improves reference fidelity through reprojection-guided test-time optimization.
We also introduce the WorldBridge benchmark suite to evaluate 3D world generation from scattered, low-overlap references, which is rarely explored by current benchmarks.
Both qualitative and quantitative results demonstrate that our method enables controllable and coherent
3D world generation, surpassing the existing state-of-the-art solutions.
Our project page is available at
\href{https://davidyan2001.github.io/WorldBridge/}{WorldBridge}.
\end{abstract}
\section{Introduction}
\label{sec:intro}

Creating explorable 3D worlds from images~\cite{Yu2024WonderWorldI3,huang2025voyager,Sun2025WorldPlayTL}
 has been a long-standing challenge in computer vision, with broad applications in robotics~\cite{Li2021iGibson2O}, AR/VR~\cite{Klein2007ParallelTA}, and video games~\cite{shaker2016procedural}.
In practical scenarios, users may wish to create a 3D world from only a few reference images, which can depict different parts of a scene, exhibit substantially different visual appearances, and share little or no visual overlap.
Such sparse observations leave large portions of a scene unobserved, making its geometry and appearance underconstrained~\cite{Niemeyer2021Regnerf,Wu2023ReconFusion3R}.
Rather than treating such ambiguity solely as a reconstruction challenge, we view it as an opportunity for controllable world generation, where users can specify how scattered references are spatially arranged while allowing unobserved regions to be plausibly synthesized.
Yet, achieving this goal requires integrating scattered visual references, user-specified constraints, and generated content into a spatially coherent 3D world.

Existing methods have made considerable progress in 3D world generation.
Image- and video-based approaches~\cite{chen2026anyrecon,zhou2025stable,gao2024cat3d,yu2024viewcrafter} use powerful generative models to synthesize realistic views 
under camera control.
However, these approaches may produce inconsistent 3D geometry across the generated views.
Geometry-aware generative methods~\cite{Lu2026GAELA,jang2026gld,Huang2026Gen3R3S,Gao2026PixWorldU3} incorporate geometric information to improve the consistency of generated appearance and scene structure.
Other approaches~\cite{ren2025gen3c,zhang2026worldstereo} maintain geometric memories to preserve information from previously observed or generated views.
Despite these advances, existing methods still struggle when the reference images have little or no visual overlap.
Geometric information from isolated observations provides limited guidance for connecting distant regions, while generative models may introduce content that does not faithfully follow the input references.
Consequently, maintaining both global scene coherence and reference fidelity remains difficult, particularly under flexible camera and text conditioning.

In this work, we present \textbf{WorldBridge}, a generative framework for creating coherent and controllable 3D worlds from scattered reference images with little or no visual overlap.
We aim to preserve the observed content while plausibly generating the unseen regions under flexible camera and text conditioning.
We coordinate shared global scene context with information from individual reference images.
The global context ensures the consistent generation of geometry and appearance across different views.
The individual references provide constraints to preserve the observed content.
This allows us to generate plausible 3D worlds while maintaining consistency with the observations.

Specifically, WorldBridge encodes the posed reference images into a shared global scene representation.
A camera-conditioned scene reader uses this representation to guide the generation of each target view.
In parallel, a separate reference encoder preserves information from individual reference images.
Our \ourattn{} mechanism further integrates temporal references along the target camera trajectory and spatial references outside the trajectory.
This mechanism allows both types of references to be used without assigning artificial temporal positions to spatial references.
A plug-and-play latent-reference consistency strategy further improves reference fidelity through reprojection-guided test-time optimization.
Together, these designs enable us to generate both RGB images and 3D geometry for the requested camera viewpoints.

Fig.~\ref{fig:teaser} illustrates the capabilities of WorldBridge under different camera and text conditioning.
First, WorldBridge can generate a plausible 3D scene from scattered views with little overlap; see Fig.~\ref{fig:teaser}(a).
As illuminated in Fig.~\ref{fig:teaser}(b), our method can also generate different spatial layouts from the same reference images by changing the camera poses.
As Fig.~\ref{fig:teaser}(c) shows, we further allow users to control unseen regions through text prompts while preserving the observed content.
Moreover, as shown in Fig.~\ref{fig:teaser}(d), our method generalizes well to references with markedly different visual styles, generating coherent 3D worlds even from combinations of real-world photographs and stylized images.

To comprehensively verify our effectiveness around 3D world generation with scattered references, we introduce the WorldBridge benchmark suite.
Existing benchmarks~\cite{zhou2018stereo, ling2024dl3dv} focus on evaluating generation on overlapping reference views, while our benchmark provides up to six scattered references with little or no visual overlap.
Specifically, WorldBridge-Short evaluates RGB synthesis from two low-overlap references in short real-world sequences of small scenes.
WorldBridge-Long evaluates visual quality, geometric consistency, camera control, and reference fidelity using multiple low-overlap references in large real-world scenes.
WorldBridge-Agent extends these evaluations to synthetic scenes, assessing the integration of both temporal references along the target trajectory and spatial references outside it.
Extensive experiments on existing novel-view-synthesis benchmarks and our proposed WorldBridge benchmark demonstrate superior view fidelity, geometric accuracy, camera controllability, consistency, and generation quality, compared with previous approaches.

To summarize, our contributions are listed as follows:
\begin{itemize}
    \item We introduce \textbf{WorldBridge}, a framework for coherent and controllable 3D world generation from scattered images with little or no visual overlap.
    It preserves observations while plausibly generating unseen regions for 3D worlds through flexible camera and text conditioning.

    \item We propose a coordinated generation strategy that combines global scene context with reference-specific information. Global-context-grounded generation provides shared guidance across views, while \ourattn{} integrates temporal and spatial
    references without requiring them to share temporal positions. A plug-and-play latent-reference consistency strategy further improves reference fidelity.

    \item We introduce the WorldBridge benchmark suite, comprising short, long, and agent-based settings, assessing reference fidelity, geometric consistency, and camera control under low-overlap and on- and off-trajectory conditions of 3D world generation.

\end{itemize}

\section{Related Work}
\label{sec:rw}

\begin{figure*}[t]
    \centering
    \includegraphics[width=.95\linewidth]
{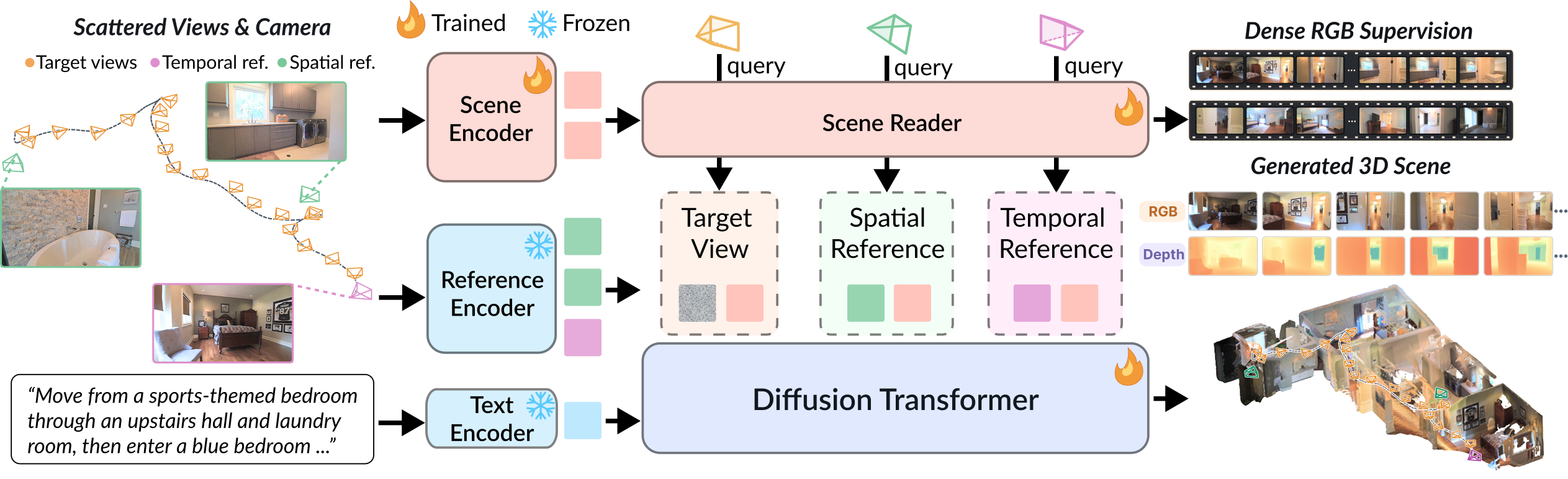}
    \caption{\textbf{Overview of WorldBridge.} The scene encoder aggregates scattered references into a global representation, which the scene reader aligns to each view under RGB supervision. The diffusion transformer combines these features with encoded temporal and spatial references and text to generate RGB and geometry.}
    \label{fig:pipeline}
\end{figure*}

\noindent\textbf{Novel View Synthesis and 3D Reconstruction.}
Novel view synthesis (NVS) generates images from unseen viewpoints
using reference observations.
Optimization-based methods~\cite{mildenhall2020nerf,Kerbl20233DGS,barron2022mipnerf,mueller2022instantngp}
reconstruct implicit or explicit 3D representations from posed images
for photorealistic rendering.
Sparse-view methods~\cite{Jain2021PuttingNO,kangle2021dsnerf,Niemeyer2021Regnerf,Yang2023FreeNeRF,wang2022sparsenerf,Zhu2023FSGSRF}
use regularization to improve reconstruction from limited observations.
Feed-forward methods~\cite{yu2021pixelnerf,wang2021ibrnet,chen2021mvsnerf,charatan2024pixelsplat,chen2024mvsplat,xu2024depthsplat,ye2024noposplat,dust3r_cvpr24,Leroy2024GroundingIM,wang2025vggt,wang2025pi,szymanowicz2026lagernvs}
further leverage learned multi-view and geometric priors to recover
3D representations from sparse or unposed images, with motion and structure priors also supporting robust monocular depth estimation under adverse conditions~\cite{Yan_2025_CVPR}.
Generative methods synthesize plausible content beyond the observed regions, drawing on image and video diffusion priors for novel view synthesis~\cite{Wu2023ReconFusion3R,gao2024cat3d,yu2024viewcrafter,li2025nvcomposer,ren2025gen3c,Wu2025GenFusion,zhang2026worldstereo,chen2026anyrecon,zhou2025stable, chen2026vggtdiff, ren2026geoverse} and learned geometric priors for 3D shape completion~\cite{Yan_2026_CVPR}.
Recent geometry-aware methods~\cite{Guizilini2025ZeroShotNV,jang2026gld,Lu2026GAELA,Huang2026Gen3R3S,wei20263drae,Gao2026PixWorldU3, kang2026geonvs,Xu2026Proxy2WorldLT}
incorporate geometric representations or model appearance and geometry jointly to improve
3D consistency, with related cross-modal alignment challenges arising in joint audio--video generation~\cite{xu2026av}.
A concurrent work \cite{lang2026roge} leverages camera-queried implicit representations for end-to-end novel view synthesis. In contrast, our work focuses on the more unconstrained setting of bridging scattered on- and off-trajectory references with little or no visual overlap to generate coherent 3D worlds.

\vspace{0.2cm}
\noindent \textbf{3D Scene Generation. }
Research on 3D content spans spatiotemporal reasoning over dynamic point clouds~\cite{zhang20264dpc} and the generation of diverse, explorable environments.
Early methods~\cite{chen2023SceneDreamer,Lin2023InfiniCityIC,Yang2023UrbanGIRAFFERU}
use learned scene representations and compositional priors to
generate large-scale environments.
Other methods~\cite{zhang2023text2nerf,Chung2023LucidDreamerDG,shriram2024realmdreamer,li2024dreamscene}
leverage pretrained image generators for text- and image-conditioned
scene synthesis.
Recent approaches~\cite{yu2023wonderjourney,Yu2024WonderWorldI3,hunyuanworld2025tencent,Wang2025WorldGenFT, Lab2026ABot3DWorld0A, feng2025wonderverse, liu2026citygenagent, wu2026phymix}
support scene expansion and interactive exploration, with multi-agent workflows also used for scientific video understanding~\cite{xu2026scieducator} and for spatially consistent video generation across clips using shared 3D layouts~\cite{shi2026one}.
Camera-controlled methods~\cite{huang2025voyager,Sun2025WorldPlayTL, shen2026lyra, dai2025fantasyworld}
further enable exploration along specified camera trajectories.
While these approaches advance controllable world generation,
integrating multiple scattered observations is not their primary focus.
WorldBridge instead generates coherent 3D worlds from multiple posed
references with little or no overlap, including references outside
the target camera trajectory.
It coordinates shared scene context with individual reference
information to guide generation while preserving observed content.
\section{Proposed Method}
\label{sec:method}

\subsection{Preliminaries}

\noindent\textbf{Geometry Autoencoder.}
WorldBridge builds on the geometry autoencoder (GAE)~\cite{Lu2026GAELA},
which represents appearance and geometry in a shared latent space.
GAE compresses multiview image features extracted by a geometric
foundation model, such as DepthAnything v3~\cite{depthanything3},
into a compact latent representation.
This representation can be decoded into RGB images and geometry
through two separate paths.
The RGB decoder predicts images directly, while the geometry decoder
and its prediction head estimate depth maps and camera rays.
We denote the encoder and two decoding paths by
$\mathcal{E}$, $\mathcal{D}_{\mathrm{rgb}}$, and
$\mathcal{D}_{\mathrm{geo}}$, respectively.
Given input views $\mathbf{I}$, the encoding and decoding processes are
\begin{equation}
\begin{gathered}
\mathbf{z} = \mathcal{E}\!\left(\mathcal{F}_{\mathrm{geo}}(\mathbf{I})\right),\\
\widehat{\mathbf{I}} = \mathcal{D}_{\mathrm{rgb}}(\mathbf{z}),\quad
(\widehat{\mathbf{d}},\widehat{\mathbf{r}})
= \mathcal{D}_{\mathrm{geo}}(\mathbf{z}),
\end{gathered}
\label{eq:gae_encoding_decoding}
\end{equation}
where $\mathcal{F}_{\mathrm{geo}}$ denotes the feature extractor of
the geometric foundation model, $\mathbf{z}$ is the shared latent
representation, and $\widehat{\mathbf{d}}$ and
$\widehat{\mathbf{r}}$ are the predicted depth maps and camera rays.

\vspace{0.3cm}
\noindent\textbf{Conditional Flow Matching.}
We train a conditional flow-matching model~\cite{Lipman2022FlowMF}
in the shared latent space:
\begin{equation}
\mathcal{L}_{\mathrm{FM}}
=
\mathbb{E}_{\mathbf{z}_0,\boldsymbol{\epsilon},t}
\!\left[
w(t)
\left\|
v_\theta(\mathbf{z}_t,t,c)
-
(\boldsymbol{\epsilon}-\mathbf{z}_0)
\right\|_2^2
\right],
\label{eq:wb_fm}
\end{equation}
where $\mathbf{z}_0$ is the clean latent encoded from the target views, and $\boldsymbol{\epsilon}\sim\mathcal{N}(0,I)$ is Gaussian noise.
The interpolated latent is
$\mathbf{z}_t=(1-t)\mathbf{z}_0+t\boldsymbol{\epsilon}$.
The model $v_\theta$ predicts the corresponding velocity
$\boldsymbol{\epsilon}-\mathbf{z}_0$, while $w(t)$ is a timestep-dependent weight.
The conditioning $c$ includes information from the reference images, target camera poses, and text guidance, as described in the following sections.

\subsection{WorldBridge Overview}

Let $\mathcal{R}=\{(I_i,P_i)\}_{i=1}^{R}$ denote the posed reference
images and $\mathcal{P}=(P_j)_{j=1}^{N}$ denote the target cameras
along an exploration trajectory.
We divide the references into two groups.
Temporal references $\mathcal{R}_{\mathrm{T}}$ have known positions
along the target sequence, while spatial references
$\mathcal{R}_{\mathrm{S}}$ have no assigned temporal positions,
such as observations outside the target trajectory.
Together, they form $\mathcal{R}=
\mathcal{R}_{\mathrm{T}}\cup\mathcal{R}_{\mathrm{S}}$.
Given $\mathcal{R}$, $\mathcal{P}$, and an optional text caption $y$,
WorldBridge generates an RGB image $\widehat I_j$ and a corresponding
3D point map $\widehat X_j$ for each target viewpoint.

As shown in Fig.~\ref{fig:pipeline}, scattered references may share
little or no visual overlap, making it difficult to maintain
coherent geometry and appearance across the generated views.
To address this problem, WorldBridge combines a shared scene
representation with separate representations of individual
references (Sec.~\ref{sec:wb_global}).
Specifically, a scene encoder $E_{\mathrm{sc}}$ aggregates the posed
references into global scene tokens $S$.
In parallel, a reference encoder $E_{\mathrm{ref}}$ uses the
pretrained GAE encoder to obtain individual reference latents
$z_i^{\mathrm{ref}}$.
To provide global information for each viewpoint,
a camera-conditioned scene reader $R_{\mathrm{sc}}$ queries
$S$ using the camera rays $r(P_i)$.
This produces a view-specific global condition $C_i$
that guides the generation of geometry and appearance
(Sec.~\ref{sec:wb_global}).
Our \ourattn{} mechanism further integrates temporal and spatial
references without assigning artificial temporal positions
to spatial references (Sec.~\ref{sec:wb_attention}).
The conditional generative transformer uses the noisy target
latents $z_{t,j}$, reference latents $z_i^{\mathrm{ref}}$,
global conditions $C_i$, and optional caption $y$
to predict the flow velocity.
The predicted velocity is used to estimate clean target latents
$\widehat z_{0,j}$ during generation.
The pretrained GAE decoders then recover RGB images, depth maps,
and camera rays, from which the corresponding 3D point maps
are obtained.

To further improve reference fidelity, a plug-and-play \ourtta{}
strategy refines the predicted clean latents during inference.
It minimizes the differences between generated views and
reference observations reprojected into the target cameras,
while keeping the model parameters fixed
(Sec.~\ref{sec:wb_tta}).

\subsection{Global-context-grounded Generation}
\label{sec:wb_global}

When scattered reference views share little or no visual overlap, cross-view feature correspondences break down completely.
Relying solely on fine-grained reference latents leaves the diffusion transformer blind to the global spatial layout--it lacks a unified structural representation to infer how disconnected regions connect across unobserved blind spots. 
To resolve the inherent tension between global structural coherence and local observation fidelity, we decouple scene conditioning into two complementary pathways: 
a global context path that infers a coarse, scene-wide spatial topology, and a local reference path that encodes each reference's appearance independently.

\vspace{0.3cm}\noindent\textbf{Dual-path Scene and Reference Encoding. }
The scene encoder $E_{\mathrm{sc}}$ maps the posed references into
shared scene tokens $S$.
In parallel, the reference encoder $E_{\mathrm{ref}}$ independently
encodes each reference image into a clean latent
$z_i^{\mathrm{ref}}$:
\begin{equation}
S=E_{\mathrm{sc}}(\mathcal{R}),\qquad
z_i^{\mathrm{ref}}=E_{\mathrm{ref}}(I_i).
\label{eq:wb_encoders}
\end{equation}
The reference encoder uses the pretrained GAE encoding path and
remains fixed during training, while the scene path is fine-tuned.
The two paths provide complementary information.
The shared tokens $S$ aggregate scene-level information across
references, while $z_i^{\mathrm{ref}}$ retains a separate
fine-grained representation of each observation.

\noindent\textbf{Camera-conditioned Scene Reading.}
The shared scene representation must provide information relevant to each requested viewpoint.
For a view $i$ with camera $P_i$, let $r(P_i)$ denote its per-patch Pl\"ucker rays.
The scene reader $R_{\mathrm{sc}}$ queries $S$ with $r(P_i)$ to obtain a view-specific condition:
\begin{equation}
C_i=W_{\mathrm{sc}}\,\mathrm{LN}
\bigl(\mathrm{Resize}(R_{\mathrm{sc}}(S,r(P_i)))\bigr).
\label{eq:wb_scene_reader}
\end{equation}
Here, $\mathrm{Resize}$ aligns the scene features with the
latent patch resolution, $\mathrm{LN}$ denotes layer normalization,
and $W_{\mathrm{sc}}$ is a projection layer.
The resulting condition is concatenated to the view tokens with the following patchify:
\begin{equation}
H_i=\mathrm{PatchProj}(z_i, C_i),
\label{eq:wb_tokens}
\end{equation}
where $z_i$ denotes the noisy target latent or clean reference
latent, $\mathrm{PatchProj}$ is the patch projection layer.
The resulting tokens are concatenated and passed to the
diffusion transformer.
In this way, each view receives guidance from the shared scene
context while retaining its own latent representation, leading to a coherent multi-view representation.

\begin{figure}[t]
    \centering
    \includegraphics[width=.95\linewidth]
    {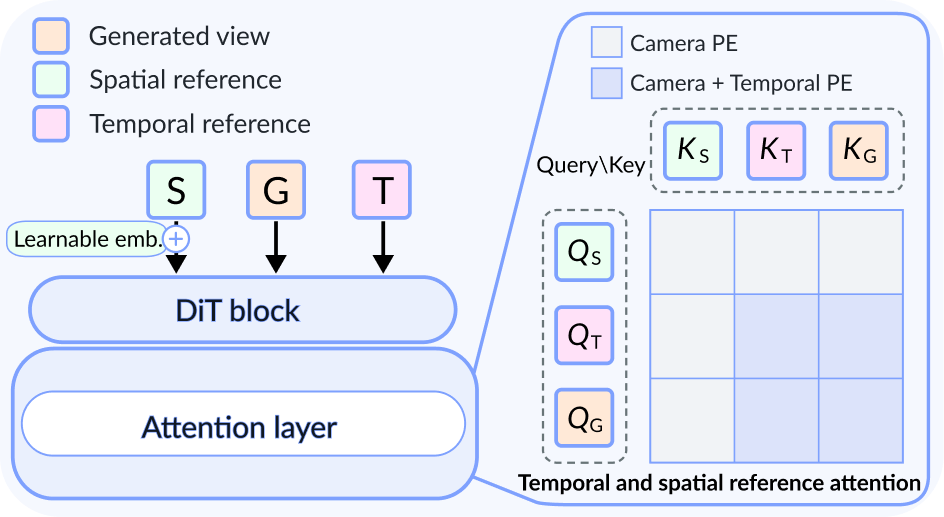}
    \caption{\textbf{Temporal and spatial reference attention.} All view pairs use camera encoding; temporal RoPE is omitted for interactions involving spatial references.}
    \label{fig:attention}
\end{figure}
\begin{figure*}[tb]
    \centering
    \includegraphics[width=0.95\linewidth]
{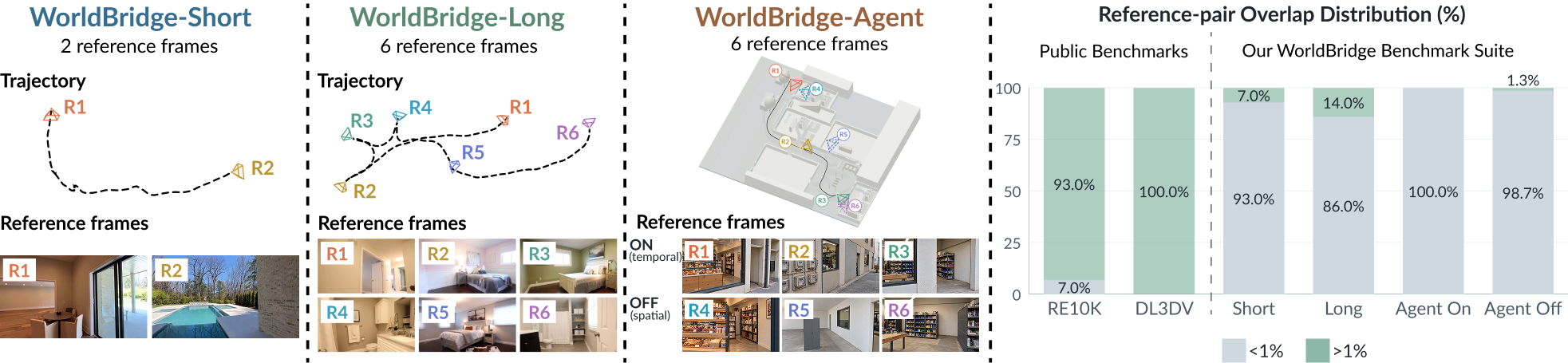}
    \caption{\textbf{Overview of the WorldBridge benchmark.} Short evaluates two-view bridging; Long evaluates long-range generation from six references; Agent uses synthetic scenes with on- and off-trajectory references. Right: WorldBridge shows a much higher fraction of reference pairs with less than 1\% overlap than the public benchmarks, evaluating world generation with low-overlapping references.
    \label{fig:benchmark}} 
\end{figure*}

\vspace{0.3cm}\noindent\textbf{Global Context Supervision.}
To encourage the shared scene representation to provide useful
information for generation, we supervise the scene reader with
low-resolution RGB predictions.
Specifically, the view-conditioned scene features are decoded
into an RGB image and compared with the corresponding training image.
Let $\mathcal{Q}$ denote the supervised camera--image pairs,
the scene supervision objective is
\begin{equation}
\mathcal{L}_{\mathrm{sc}}
=
\sum_{(P,I)\in\mathcal{Q}}
\ell_{\mathrm{rgb}}
\bigl(\widehat I_{\mathrm{sc}}(S,P),I\bigr),
\label{eq:wb_scene_loss}
\end{equation}
where $\widehat I_{\mathrm{sc}}(S,P)$ is the predicted image
and $\ell_{\mathrm{rgb}}$ denotes the mean squared error.
This supervision encourages the shared scene representation to
provide view-relevant appearance and structural information.
The overall training objective combines scene supervision and flow matching loss:
\begin{equation}
\mathcal{L}=\mathcal{L}_{\mathrm{FM}}+\mathcal{L}_{\mathrm{sc}}.
\label{eq:wb_total_loss}
\end{equation}

\subsection{Temporal and Spatial Reference Attention}
\label{sec:wb_attention}

As shown in Fig.~\ref{fig:attention}, WorldBridge integrates
temporal references $\mathcal{R}_{\mathrm{T}}$, spatial references
$\mathcal{R}_{\mathrm{S}}$, and generated target views $\mathcal{G}$
through joint attention.
Temporal references have known positions in the target sequence,
while spatial references have camera poses but no assigned
sequence positions.
To avoid introducing artificial temporal relationships, we apply
selective temporal positional encoding and use learnable role
embeddings for spatial references.

During training, temporal references are sampled from the target
frames, while spatial references are sampled from other frames
of the same video.
We also randomly treat some temporal references as spatial
references, retaining their camera poses but omitting their
temporal positions.

Specifically, temporal RoPE is applied only when both tokens
in an attention pair have known sequence positions:
\begin{equation}
m_{ij}=\mathbf{1}
[i\in\mathcal{G}\cup\mathcal{R}_{\mathrm{T}}
\land j\in\mathcal{G}\cup\mathcal{R}_{\mathrm{T}}].
\label{eq:wb_queries}
\end{equation}
For pairs involving spatial references ($m_{ij}=0$),
temporal RoPE is omitted.
Spatial RoPE and the SCoPE~\cite{yin2026scope} camera encoding
based on Pl\"ucker rays are retained for all pairs.
This allows WorldBridge to integrate temporal and spatial
references without imposing artificial temporal positions.

\subsection{Latent-Reference Consistency}
\label{sec:wb_tta}

\begin{table*}[t]
\centering
\setlength{\belowcaptionskip}{3pt}
\caption{Quantitative comparison of novel-view synthesis on public benchmarks: RealEstate10K~\citep{zhou2018stereo} and DL3DV-10K~\citep{ling2024dl3dv} under 1-view and 2-view input settings, following the evaluation split of \citep{celen2026genrec}. Best results are in \textbf{bold} and second best are \underline{underlined}.}
\label{tab:nvs_comparison}
\renewcommand{\arraystretch}{1.0}
\setlength{\aboverulesep}{0.25ex}
\setlength{\belowrulesep}{0.35ex}
\footnotesize
\setlength{\tabcolsep}{1.5pt}
\newcommand{\wbOneHeader}[1]{{\scriptsize #1}}
\begin{tabular*}{\textwidth}{@{\extracolsep{\fill}}l*{12}{c}@{}}
\toprule
& \multicolumn{6}{c}{\textbf{RealEstate10K}} & \multicolumn{6}{c}{\textbf{DL3DV-10K}} \\
\cmidrule(lr){2-7} \cmidrule(lr){8-13}
& \multicolumn{3}{c}{1-view} & \multicolumn{3}{c}{2-view} & \multicolumn{3}{c}{1-view} & \multicolumn{3}{c}{2-view} \\
\cmidrule(lr){2-4} \cmidrule(lr){5-7} \cmidrule(lr){8-10} \cmidrule(lr){11-13}\makebox[62pt][l]{Method}\makebox[42pt][l]{Venue}
& \wbOneHeader{PSNR$\uparrow$} & \wbOneHeader{SSIM$\uparrow$} & \wbOneHeader{LPIPS$\downarrow$}
& \wbOneHeader{PSNR$\uparrow$} & \wbOneHeader{SSIM$\uparrow$} & \wbOneHeader{LPIPS$\downarrow$}
& \wbOneHeader{PSNR$\uparrow$} & \wbOneHeader{SSIM$\uparrow$} & \wbOneHeader{LPIPS$\downarrow$}
& \wbOneHeader{PSNR$\uparrow$} & \wbOneHeader{SSIM$\uparrow$} & \wbOneHeader{LPIPS$\downarrow$} \\
\midrule
\makebox[62pt][l]{SEVA~\citep{zhou2025stable}}\makebox[42pt][l]{\scriptsize ICCV'25}
& 13.48 & 0.453 & 0.493
& 18.47 & \underline{0.648} & \underline{0.271}
& 13.11 & 0.321 & 0.520
& \underline{18.84} & \underline{0.570} & \underline{0.251} \\
\makebox[62pt][l]{GAE~\citep{Lu2026GAELA}}\makebox[42pt][l]{\scriptsize arXiv'26}
& 15.77 & \underline{0.557} & \underline{0.378}
& 18.06 & 0.637 & 0.293
& \underline{16.63} & \underline{0.472} & \underline{0.351}
& 18.63 & 0.553 & 0.268 \\
\makebox[62pt][l]{GLD~\citep{jang2026gld}}\makebox[42pt][l]{\scriptsize ECCV'26}
& 14.71 & 0.505 & 0.462
& \underline{18.59} & 0.634 & 0.300
& 15.27 & 0.388 & 0.443
& 18.76 & 0.521 & 0.298 \\
\makebox[62pt][l]{Gen3R~\citep{Huang2026Gen3R3S}}\makebox[42pt][l]{\scriptsize CVPR'26}
& 13.80 & 0.461 & 0.490
& 16.93 & 0.573 & 0.342
& 14.75 & 0.365 & 0.423
& 17.33 & 0.470 & 0.297 \\
\makebox[62pt][l]{AnyRecon~\citep{chen2026anyrecon}}\makebox[42pt][l]{\scriptsize SIGA'26}
& \underline{16.08} & 0.535 & 0.386
& 17.97 & 0.605 & 0.295
& 15.88 & 0.406 & 0.360
& 17.70 & 0.492 & 0.259 \\
\makebox[62pt][l]{3DRAE~\citep{wei20263drae}}\makebox[42pt][l]{\scriptsize NeurIPS'26}
& 14.10 & 0.469 & 0.491
& 16.83 & 0.549 & 0.361
& 14.29 & 0.343 & 0.489
& 16.26 & 0.412 & 0.397 \\
\midrule
Ours & \textbf{16.71} & \textbf{0.601} & \textbf{0.351} & \textbf{20.88} & \textbf{0.756} & \textbf{0.219} & \textbf{17.48} & \textbf{0.505} & \textbf{0.336} & \textbf{21.33} & \textbf{0.690} & \textbf{0.205} \\
\bottomrule
\end{tabular*}
\end{table*}
\begin{table*}[t]
\centering
\setlength{\belowcaptionskip}{3pt}
\caption{Quantitative comparison on \textbf{WorldBridge-Short}. Best results are in \textbf{bold} and second best are \underline{underlined}.}
\label{tab:worldbridge100}
\renewcommand{\arraystretch}{1.0}
\setlength{\aboverulesep}{0.25ex}
\setlength{\belowrulesep}{0.35ex}
\footnotesize
\setlength{\tabcolsep}{1.5pt}
\newcommand{\wbTwoHeader}[1]{{\scriptsize #1}}
\begin{tabular*}{\textwidth}{@{\extracolsep{\fill}}l*{12}{c}@{}}
\toprule
&\multicolumn{3}{c}{\textbf{RealEstate10K}}&\multicolumn{3}{c}{\textbf{DL3DV-10K}}&\multicolumn{3}{c}{\textbf{SpatialVID}}&\multicolumn{3}{c}{\textbf{Overall}}\\
\cmidrule(lr){2-4}\cmidrule(lr){5-7}\cmidrule(lr){8-10}\cmidrule(lr){11-13}
\makebox[62pt][l]{Method}\makebox[42pt][l]{Venue}
&\wbTwoHeader{PSNR$\uparrow$}&\wbTwoHeader{SSIM$\uparrow$}&\wbTwoHeader{LPIPS$\downarrow$}
&\wbTwoHeader{PSNR$\uparrow$}&\wbTwoHeader{SSIM$\uparrow$}&\wbTwoHeader{LPIPS$\downarrow$}
&\wbTwoHeader{PSNR$\uparrow$}&\wbTwoHeader{SSIM$\uparrow$}&\wbTwoHeader{LPIPS$\downarrow$}
&\wbTwoHeader{PSNR$\uparrow$}&\wbTwoHeader{SSIM$\uparrow$}&\wbTwoHeader{LPIPS$\downarrow$}\\
\midrule
\makebox[62pt][l]{SEVA~\citep{zhou2025stable}}\makebox[42pt][l]{\scriptsize ICCV'25}
& 15.36 & 0.542 & \underline{0.380}
& 15.71 & 0.423 & \underline{0.456}
& 15.88 & 0.462 & \underline{0.417}
& 15.66 & 0.471 & \underline{0.421} \\
\makebox[62pt][l]{GAE~\citep{Lu2026GAELA}}\makebox[42pt][l]{\scriptsize arXiv'26}
& 15.68 & 0.552 & 0.387
& 15.23 & 0.413 & 0.485
& 15.02 & 0.443 & 0.466
& 15.29 & 0.464 & 0.450 \\
\makebox[62pt][l]{GLD~\citep{jang2026gld}}\makebox[42pt][l]{\scriptsize ECCV'26}
& \underline{15.97} & \underline{0.559} & 0.389
& \underline{16.47} & \underline{0.454} & 0.458
& \underline{15.90} & \underline{0.476} & 0.429
& \underline{16.13} & \underline{0.492} & 0.428 \\

\makebox[62pt][l]{Gen3R~\citep{Huang2026Gen3R3S}}\makebox[42pt][l]{\scriptsize CVPR'26}
& 14.21 & 0.486 & 0.461
& 14.59 & 0.383 & 0.562
& 15.29 & 0.430 & 0.464
& 14.72 & 0.429 & 0.499 \\
\makebox[62pt][l]{AnyRecon~\citep{chen2026anyrecon}}\makebox[42pt][l]{\scriptsize SIGA'26}
& 14.94 & 0.512 & 0.412
& 15.45 & 0.403 & 0.481
& 13.84 & 0.393 & 0.490
& 14.76 & 0.431 & 0.464 \\
\makebox[62pt][l]{3DRAE~\citep{wei20263drae}}\makebox[42pt][l]{\scriptsize NeurIPS'26}
& 14.92 & 0.511 & 0.440
& 15.96 & 0.415 & 0.489
& 15.01 & 0.428 & 0.478
& 15.34 & 0.447 & 0.471 \\
\midrule
Ours & \textbf{18.17} & \textbf{0.676} & \textbf{0.279} & \textbf{18.22} & \textbf{0.540} & \textbf{0.385} & \textbf{18.65} & \textbf{0.613} & \textbf{0.322} & \textbf{18.35} & \textbf{0.604} & \textbf{0.333} \\
\bottomrule
\end{tabular*}
\end{table*}
\begin{table*}[t]
\centering
\setlength{\belowcaptionskip}{3pt}
\caption{Quantitative comparison on \textbf{WorldBridge-Long}. \textemdash~ means that the corresponding
method does not 
provide
native geometry output.
Best results are in \textbf{bold} and second best are \underline{underlined}.}
\label{tab:worldbridge_extended}
\begingroup
\footnotesize
\setlength{\tabcolsep}{1.5pt}
\renewcommand{\arraystretch}{1.0}
\setlength{\aboverulesep}{0.25ex}
\setlength{\belowrulesep}{0.35ex}
\newcommand{\wbFourHeader}[1]{{\scriptsize #1}}
\begin{tabular*}{\textwidth}{@{\extracolsep{\fill}}l*{12}{c}@{}}
\toprule
& \multicolumn{3}{c}{\textbf{RGB fidelity}}
& \multicolumn{2}{c}{\textbf{Native geo.}}
& \multicolumn{3}{c}{\textbf{Camera control}}
& \multicolumn{2}{c}{\textbf{Consistency}}
& \textbf{Ref.}
& \textbf{Gen. quality} \\
\cmidrule(lr){2-4}\cmidrule(lr){5-6}\cmidrule(lr){7-9}\cmidrule(lr){10-11}\cmidrule(lr){12-12}\cmidrule(l){13-13}
\makebox[62pt][l]{Method}\makebox[42pt][l]{Venue}
& \wbFourHeader{PSNR$\uparrow$}
& \wbFourHeader{SSIM$\uparrow$}
& \wbFourHeader{LPIPS$\downarrow$}
& \wbFourHeader{AbsRel$\downarrow$}
& \wbFourHeader{$\delta_1\uparrow$}
& \wbFourHeader{ATE$\downarrow$}
& \wbFourHeader{RPE$_r\downarrow$}
& \wbFourHeader{RPE$_t\downarrow$}
& \wbFourHeader{Feat. reproj.$\downarrow$}
& \wbFourHeader{Geo. reproj.$\downarrow$}
& \wbFourHeader{Ref. err.$\downarrow$}
& \wbFourHeader{Aesthetic$\uparrow$} \\
\midrule
\makebox[62pt][l]{SEVA~\citep{zhou2025stable}}\makebox[42pt][l]{\scriptsize ICCV'25}
& 15.41 & 0.499 & 0.459 & \textemdash & \textemdash
& 2.810 & 9.270 & 1.378 & 0.220 & 0.127 & 0.257 & \underline{5.228} \\
\makebox[62pt][l]{GAE~\citep{Lu2026GAELA}}\makebox[42pt][l]{\scriptsize arXiv'26}& 12.86 & 0.405 & 0.573 & 0.599 & 44.9& \underline{1.508} & 10.382 & 0.717 & 0.200 & 0.110 & 0.415 & 5.143 \\
\makebox[62pt][l]{GLD~\citep{jang2026gld}}\makebox[42pt][l]{\scriptsize ECCV'26}
& \underline{16.43} & \underline{0.524} & \underline{0.436} & \underline{0.454} & \underline{59.4}
& 1.748 & 9.324 & 1.039 & 0.217 & 0.123 & \underline{0.252} & 4.920 \\
\makebox[62pt][l]{Gen3R~\citep{Huang2026Gen3R3S}}\makebox[42pt][l]{\scriptsize CVPR'26}
& 15.47 & 0.487 & 0.481 & 0.507 & 41.6
& 1.675 & \underline{8.199} & \underline{0.559} & 0.196 & 0.110 & 0.305 & 4.900 \\
\makebox[62pt][l]{AnyRecon~\citep{chen2026anyrecon}}\makebox[42pt][l]{\scriptsize SIGA'26}
& 15.87 & 0.508 & 0.444 & \textemdash & \textemdash
& 4.539 & 11.244 & 1.439 & 0.211 & 0.128 & 0.277 & 4.949 \\
\makebox[62pt][l]{3DRAE~\citep{wei20263drae}}\makebox[42pt][l]{\scriptsize NeurIPS'26}
& 15.38 & 0.469 & 0.489 & 0.475 & 53.1
& 2.427 & 11.772 & 0.995 & \underline{0.176} & \underline{0.106} & 0.332 & 4.736 \\
\midrule
Ours & \textbf{17.68} & \textbf{0.587} & \textbf{0.384} & \textbf{0.345} & \textbf{71.0} & \textbf{0.657} & \textbf{3.456} & \textbf{0.397} & \textbf{0.173} & \textbf{0.099} & \textbf{0.200} & \textbf{5.248} \\
\bottomrule
\end{tabular*}
\endgroup
\end{table*}
\begin{table*}[t]
\centering
\begin{minipage}{\textwidth}
\centering
\setlength{\belowcaptionskip}{3pt}
\caption{Quantitative comparison on \textbf{WorldBridge-Agent}.
Each entry reports On / Off trajectory results and averages worlds equally.
Best results are in \textbf{bold} and second best are \underline{underlined}.}
\label{tab:worldbridge_agent}
\begingroup
\footnotesize
\setlength{\tabcolsep}{4pt}
\renewcommand{\arraystretch}{1.05}
\setlength{\aboverulesep}{0.25ex}
\setlength{\belowrulesep}{0.35ex}
\newcommand{\wbFivePending}{\textcolor{black!45}{TBD}}
\newcommand{\wbFivePair}[2]{\mbox{#1\kern0.32em/\kern0.32em#2}}
\newcommand{\wbFiveHeader}[1]{{\scriptsize #1}}
\begin{tabular*}{\linewidth}{@{\extracolsep{\fill}}l*{7}{c}@{}}
\toprule
& \multicolumn{3}{c}{\textbf{Camera control}}
& \multicolumn{2}{c}{\textbf{Consistency}}
& \textbf{Ref. fidelity}
& \textbf{Gen. quality} \\
\cmidrule(lr){2-4}\cmidrule(lr){5-6}\cmidrule(lr){7-7}\cmidrule(l){8-8}
\makebox[62pt][l]{Method}\makebox[42pt][l]{Venue}
& \wbFiveHeader{ATE$\downarrow$}
& \wbFiveHeader{RPE$_r\downarrow$}
& \wbFiveHeader{RPE$_t\downarrow$}
& \wbFiveHeader{Feat. reproj.$\downarrow$}
& \wbFiveHeader{Geo. reproj.$\downarrow$}
& \wbFiveHeader{Ref. err.$\downarrow$}
& \wbFiveHeader{Aesthetic$\uparrow$} \\
\midrule
\makebox[62pt][l]{SEVA~\citep{zhou2025stable}}\makebox[42pt][l]{\scriptsize ICCV'25}
& \wbFivePair{5.705}{5.091}
& \wbFivePair{10.467}{7.556}
& \wbFivePair{3.144}{3.529}
& \wbFivePair{0.176}{0.237}
& \wbFivePair{0.077}{0.110}
& \wbFivePair{0.298}{0.348}
& \wbFivePair{\underline{5.561}}{\underline{5.557}} \\
\makebox[62pt][l]{GAE~\citep{Lu2026GAELA}}\makebox[42pt][l]{\scriptsize arXiv'26}
& \wbFivePair{5.935}{5.928}
& \wbFivePair{14.948}{18.726}
& \wbFivePair{2.974}{2.485}
& \wbFivePair{0.217}{0.257}
& \wbFivePair{0.108}{0.134}
& \wbFivePair{0.441}{0.475}
& \wbFivePair{5.345}{5.100} \\
\makebox[62pt][l]{GLD~\citep{jang2026gld}}\makebox[42pt][l]{\scriptsize ECCV'26}
& \wbFivePair{5.680}{6.257}
& \wbFivePair{11.264}{15.224}
& \wbFivePair{2.759}{3.534}
& \wbFivePair{0.194}{0.258}
& \wbFivePair{0.085}{0.133}
& \wbFivePair{\underline{0.282}}{\textbf{0.325}}
& \wbFivePair{5.488}{5.338} \\
\makebox[62pt][l]{Gen3R~\citep{Huang2026Gen3R3S}}\makebox[42pt][l]{\scriptsize CVPR'26}
& \wbFivePair{7.786}{9.520}
& \wbFivePair{19.293}{25.855}
& \wbFivePair{3.217}{2.849}
& \wbFivePair{0.225}{0.331}
& \wbFivePair{0.100}{0.151}
& \wbFivePair{0.377}{0.435}
& \wbFivePair{4.976}{4.747} \\
\makebox[62pt][l]{AnyRecon~\citep{chen2026anyrecon}}\makebox[42pt][l]{\scriptsize SIGA'26}
& \wbFivePair{\underline{3.711}}{4.680}
& \wbFivePair{7.428}{12.919}
& \wbFivePair{\underline{2.105}}{3.665}
& \wbFivePair{0.154}{0.201}
& \wbFivePair{\underline{0.066}}{0.094}
& \wbFivePair{0.342}{0.365}
& \wbFivePair{5.422}{5.349} \\
\makebox[62pt][l]{3DRAE~\citep{wei20263drae}}\makebox[42pt][l]{\scriptsize NeurIPS'26}
& \wbFivePair{4.194}{\underline{4.576}}
& \wbFivePair{\underline{6.253}}{\underline{4.944}}
& \wbFivePair{2.117}{\underline{2.091}}
& \wbFivePair{\underline{0.145}}{\underline{0.162}}
& \wbFivePair{0.067}{\underline{0.078}}
& \wbFivePair{0.376}{0.394}
& \wbFivePair{5.169}{5.018} \\
\midrule
Ours
& \wbFivePair{\textbf{3.278}}{\textbf{3.303}}
& \wbFivePair{\textbf{6.093}}{\textbf{4.643}}
& \wbFivePair{\textbf{1.416}}{\textbf{1.331}}
& \wbFivePair{\textbf{0.143}}{\textbf{0.149}}
& \wbFivePair{\textbf{0.057}}{\textbf{0.061}}
& \wbFivePair{\textbf{0.239}}{\underline{0.344}}
& \wbFivePair{\textbf{5.648}}{\textbf{5.675}} \\
\bottomrule
\end{tabular*}
\endgroup
\end{minipage}
\end{table*}

To further preserve observed content, inspired by~\cite{yan2026lascomp}, we propose a plug-and-play strategy to refine the predicted latents during inference.
For each reference, we estimate a metric-scale 3D point map using MoGe-3~\cite{Kong2026MoGe3FM} and project its RGB points into each target camera $P_j$.
This produces a pseudo-image $\widetilde I_j$ and valid mask $M_j$.
Unreliable depth boundaries are discarded, and a depth buffer resolves projection conflicts.

At timestep $t$, the predicted clean latents $\widehat z_0$
are decoded into RGB views
$\widehat I_j=\mathcal{D}_{\mathrm{rgb}}(\widehat z_0)_j$.
We define $\mathcal{L}_{\mathrm{proj}}$ as the masked MSE between
$\widehat I_j$ and $\widetilde I_j$, averaged over target views
with valid projections.
A direct RGB loss is also applied when a target camera coincides
with a reference camera.
Keeping the model parameters fixed, we refine the clean latents
with one gradient step and a step size $\eta$:
\begin{equation}
z^{*}_0=\widehat z_0-\eta\,\nabla_{\widehat z_0}
\mathcal{L}_{\mathrm{proj}}.
\label{eq:wb_tta_update}
\end{equation}

To continue denoising from $t$ to $t'<t$, we re-noise each refined latent using estimated and fresh Gaussian noise:
\begin{equation}
z_{t',j}=(1-t')z^{*}_{0,j}
+t'\left(
\sqrt{1-\lambda_{t,j}}\,\widehat\epsilon_j
+\sqrt{\lambda_{t,j}}\,\epsilon_j
\right),
\label{eq:wb_tta_renoise}
\end{equation}
where $\widehat\epsilon_j=
(z_{t,j}-(1-t)\widehat z_{0,j})/t$ and
$\epsilon_j\sim\mathcal{N}(0,I)$.
The weight $\lambda_{t,j}\in[0,1]$ depends on the
denoising timestep and increases with the frame index distance to the nearest reference or frame nearest to the spatial reference.
This balances fidelity and flexibility in generating
unseen regions.
Refinement and re-noising are repeated in denoising.
\section{Experiments}
\label{sec:exp}

\subsection{WorldBridge Benchmark}

Existing evaluation benchmarks do not fully capture the challenges of preserving scattered observations while generating coherent scenes along extended camera trajectories, especially when references have little or even no overlap and lie outside the target path. 
To address this gap, we introduce the WorldBridge benchmark suite (\cref{fig:benchmark}), comprising three complementary settings that assess low-overlap view bridging, long-range multi-reference generation, and the integration of off-trajectory observations.
We also provide a reference overlap comparison on the right of (\cref{fig:benchmark}) between ours and public benchmarks to show the low-overlap world generation setting of our benchmark.

\paragraph{WorldBridge-Short}
evaluates view bridging between two low-overlap references. It contains
100 clips of 5--15 seconds from
RealEstate10K~\cite{zhou2018stereo},
DL3DV~\cite{ling2024dl3dv}, and
SpatialVID~\cite{wang2025spatialvid}, using the first and last frames
as references. We first screen candidate clips using Qwen3.5~\cite{qwen3.5} and RoMa v2~\cite{edstedt2025romav2}, then use ViPE~\cite{Huang2025ViPEVP} and
DepthAnything v3~\cite{depthanything3} to estimate camera poses
and depth. 
Each clip provides at most six intermediate target views,
each with at least 50\% of its pixels estimated to be visible in one
or both references. This setting primarily assesses RGB fidelity in
reference-supported regions.

\paragraph{WorldBridge-Long}
evaluates long-range generation from multiple references, containing
20 indoor and outdoor videos of 20--60 seconds from
SpatialVID~\cite{wang2025spatialvid}, each with six references and
50 non-reference target views. The references cover complementary
regions along the trajectory, with low overlap encouraged between
non-adjacent views. We use the same annotation pipeline as Short and
evaluate RGB and reference fidelity, geometry accuracy, camera control, cross-view consistency, and generation quality.

\paragraph{WorldBridge-Agent} evaluates generation from references on and off the target trajectory.
It comprises 20 synthetic indoor and outdoor scenes created through an agent-assisted workflow: an agent designs a scene blockout and camera
trajectory, which provide structural guides for reference image generation. Each scene has two six-reference settings sharing the same target trajectory, with references either on or off that trajectory.
We use the same evaluation categories as Long, except RGB fidelity and geometry accuracy, since captured target images and ground-truth geometry are unavailable. Additional benchmark details are provided in the supplementary material.

\subsection{Experimental Settings}
\paragraph{Implementation Details.} We implement our pipeline based on GAE~\cite{Lu2026GAELA}, which provides a good geometry-appearance space for scene generation. 
The scene encoder and scene reader are initialized from LagerNVS~\cite{szymanowicz2026lagernvs} and tuned; the reference and text encoders are DepthAnything v3~\cite{depthanything3} and Qwen3-0.6B~\cite{yang2025qwen3} and kept frozen; the pipeline has 0.94B trainable parameters in total.
We train our model on a collection of video clips from RealEstate10K~\cite{zhou2018stereo}, DL3DV~\cite{ling2024dl3dv}, and SpatialVID~\cite{wang2025spatialvid}, which contains about 22k video clips at $672\times378$ and is separate from the evaluation benchmarks.
The model is trained on 32 GPUs for 10 epochs using the Adam optimizer with a batch size of 64, and a learning rate decaying from $5 \times 10^{-5}$ to $5 \times 10^{-6}$.
During training, we use a mixing sample strategy to sample temporal and spatial references from the video clips.
More implementation details are in the supplementary material.

\paragraph{Baselines.}
We compare our method against a comprehensive set of recent scene generation approaches~\cite{zhou2025stable, Lu2026GAELA, jang2026gld, Huang2026Gen3R3S, chen2026anyrecon, wei20263drae}, including the view-based generative methods and the scene-query-based methods.
All generative methods are evaluated under their official settings, while ours uses 50 denoising steps and a classifier-free guidance (CFG) scale of 2.
For prior methods, we align the scale of the camera trajectory to their proper scale; for \cite{chen2026anyrecon}, we align the geometry from Pi3~\cite{wang2025pi} to the given camera's scale and do reprojection with aligned geometry and the given camera.
Differences in supported reference configurations and inference procedures are detailed in the supplementary material.

\paragraph{Evaluation Datasets and Metrics.}
We assess all methods on the public novel-view-synthesis benchmarks and our proposed WorldBridge benchmark suite. 
The public NVS benchmarks include RealEstate10K~\cite{zhou2018stereo} and DL3DV~\cite{ling2024dl3dv}, both including 100 1-view and 100 2-view samples, and we follow the evaluation split of \cite{celen2026genrec}.
Experiments on public NVS benchmarks and our WorldBridge-Short focus on novel view generation fidelity, assessing PSNR, SSIM, and LPIPS.
Experiments on WorldBridge-Long and WorldBridge-Agent target assessing camera controllability (ATE and rotational and translational RPE), scene consistency (feature and geometric reprojection errors), reference fidelity (reference-to-target reprojection LPIPS), and generation quality (aesthetic score). 
The evaluation of camera control and consistency utilizes the geometry estimated from VGGT-$\Omega$~\cite{wang2026vggtomega}.
As WorldBridge-Long contains the labels for captured RGB images and estimated depth annotations, we also assess the novel view fidelity and geometry accuracy (AbsRel, $\delta_1$).
More details are listed in the supp.

\begin{figure*}[t]
    \centering
    \includegraphics[width=.95\linewidth]
{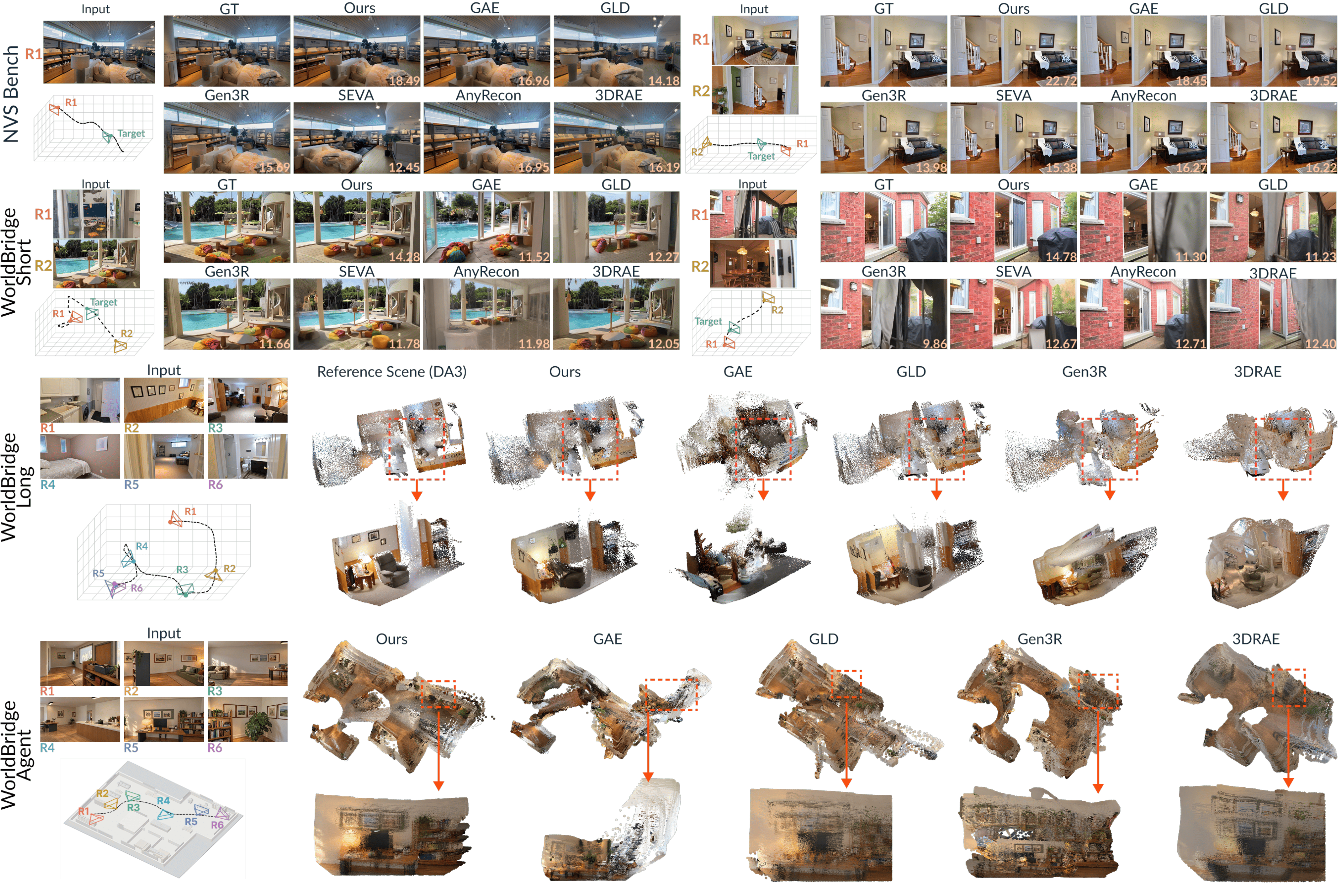}
    \caption{\textbf{Qualitative comparisons.} Top: novel views on public NVS benchmarks and WorldBridge-Short. Bottom: point clouds and local details on WorldBridge-Long/Agent.  The Long reference scene is estimated with DA3. Orange numbers indicate PSNR. 
    \label{fig:qualitative}} 
    \vspace{-2mm}
\end{figure*}
\subsection{Quantitative Results}
Tables~\ref{tab:nvs_comparison}--\ref{tab:worldbridge_agent} evaluate quantitative results on different benchmarks. Ours achieves the best RGB fidelity on RealEstate10K and DL3DV under one- and two-view settings (\cref{tab:nvs_comparison}). 
On the low-overlap WorldBridge-Short, it improves PSNR from 16.13 to 18.35 dB over the best baseline (\cref{tab:worldbridge100}), demonstrating effective sparse-reference integration. 
For long-range generation, Ours improves PSNR from 16.43 to 17.68 dB and reduces AbsRel against estimated depth from 0.454 to 0.345 compared with GLD, while achieving the lowest geometric reprojection error (\cref{tab:worldbridge_extended}). 
On WorldBridge-Agent, it achieves the lowest camera-pose and geometric reprojection errors with both on- and off-trajectory references (\cref{tab:worldbridge_agent}), supporting flexible reference conditioning.

\subsection{Qualitative Results}
As shown in \cref{fig:qualitative}, WorldBridge produces faithful novel views under different reference settings. With a single or two references, it better retains scene content across the input views, including on WorldBridge-Short where the references have little overlap.
For the multi-reference settings on WorldBridge-Long and WorldBridge-Agent, ours brings scattered observations together into coherent scenes while retaining local appearance near the references. 
The generated point clouds also exhibit clearer, more continuous scene structures than several baselines, whose results show fragmented geometry or blurred details. Together, these examples illustrate how WorldBridge combines sparse-view fidelity with scene-level geometric coherence.

\subsection{Applications}
As shown in \cref{fig:teaser}, WorldBridge reconstructs scenes from sparse references and controls their layout via reference camera poses. Text prompts guide the content of unseen regions, while combining real and stylized references enables imaginative world generation. More visualizations are provided in the supplementary material.
\section{Ablation Studies}
\label{sec:ablations}
\begin{table}[t]
\centering
\caption{Ablations on WorldBridge-Long / Agent. For ATE, geometric reprojection, and reference error, each entry reports Long / Agent On / Agent Off. PSNR is evaluated on Long only.}
\label{tab:ablation_training}
\begingroup\footnotesize\setlength{\tabcolsep}{1.5pt}
\renewcommand{\arraystretch}{1.15}
\newcommand{\wbtrip}[3]{#1/#2/\makebox[2.5em][c]{#3}}
\resizebox{\columnwidth}{!}{%
\begin{tabular}{@{}l@{\hspace{4pt}}c@{\hspace{4pt}}c@{\hspace{4pt}}c@{\hspace{4pt}}c@{}}
\toprule
& \multicolumn{4}{c}{\textbf{Long / Agent (On / Off)}} \\
\cmidrule(l){2-5}
Variant & PSNR$\uparrow$ & ATE$\downarrow$ & Geo. reproj.$\downarrow$ & Ref. err.$\downarrow$ \\
\midrule
w/o global context & 16.61 & \wbtrip{0.834}{3.731}{4.714} & \wbtrip{0.115}{0.080}{0.126} & \wbtrip{0.254}{0.299}{0.445} \\
w/o $\mathcal{L}_{\mathrm{sc}}$ & 17.18 & \wbtrip{0.933}{3.477}{3.443} & \wbtrip{0.109}{0.061}{0.080} & \wbtrip{0.233}{0.272}{0.371} \\
temporal ref. only
& 17.35 & 0.850/3.566/3.402 & 0.105/0.061/0.085 & 0.218/0.247/0.391 \\
spatial ref. only & 17.19 & \wbtrip{0.840}{3.339}{3.669} & \wbtrip{0.102}{0.056}{0.077} & \wbtrip{0.220}{0.258}{0.368} \\
w/o LRC & 17.30 & \wbtrip{0.787}{3.218}{3.415} & \wbtrip{0.101}{0.056}{0.064} & \wbtrip{0.214}{0.254}{0.361} \\
\midrule
full model & 17.68 & \wbtrip{0.657}{3.278}{3.303} & \wbtrip{0.099}{0.057}{0.061} & \wbtrip{0.200}{0.239}{0.344} \\
\bottomrule
\end{tabular}%
}
\endgroup
\end{table}

\cref{tab:ablation_training} ablates the global scene branch, its RGB supervision, the reference types used during training (\emph{temporal ref. only} means giving a fixed temporal index to all spatial references, while \emph{spatial ref. only} treats all references without temporal RoPE), and latent-reference consistency (LRC) at inference. 
Removing the global scene branch lowers PSNR on WorldBridge-Long from 17.68 to 16.61 dB and more than doubles geometric reprojection error on Agent Off, from 0.061 to 0.126. 
With the scene branch retained, removing its RGB supervision ($\mathcal{L}_{\mathrm{sc}}$) lowers Long PSNR from 17.68 to 17.18 dB and increases reference error in all three settings.
Training with both temporal and spatial references yields lower reference error than training with either type alone on Long, Agent On/Off. The difference is largest on Agent Off, where reference error is 0.344 with mixed training, compared with 0.391 with temporal references alone and 0.368 with spatial references alone.
LRC raises PSNR from 17.30 to 17.68 dB and lowers reference error in all settings, with a reduction from 0.361 to 0.344 on Agent Off. 

\vspace{-1mm}
\section{Conclusion}
\label{sec:conclusion}

We present WorldBridge, which generates coherent 3D worlds from scattered images with little or no overlap.
Global scene context guides each view, while temporal and spatial reference attention preserves observed details.
Latent-reference consistency improves reference fidelity at inference.
Our proposed benchmark suite assesses low-overlap view bridging, long-range generation, and the use of off-trajectory references.
Results on diverse benchmarks show the effectiveness of our pipeline.

\noindent\textbf{Limitations.} 
Although our method generates high-quality 3D scenes, it may lack fine-detailed and structural geometry.
When target views have little visual overlap with the spatial references, these references provide limited guidance, which may reduce generation quality.
Further discussion is provided in the supplementary material.

\twocolumn[{
    \centering
    {\Large\bfseries Supplementary Material\par}
    \vspace{0.75em}
}]
\appendix

This supplementary material describes the training and inference procedures for WorldBridge (Sec.~\ref{sec:implementation}),
the construction of the WorldBridge benchmark suite (Sec.~\ref{sec:benchmark}), 
and the evaluation settings and metrics (Sec.~\ref{sec:eval_protocol}). 
We present additional qualitative comparisons and application examples (Sec.~\ref{sec:additional_results}), 
and discuss limitations and the role of global scene context (Sec.~\ref{sec:discussions}).

\section{Implementation Details}
\label{sec:implementation}

\subsection{Training Configuration}
We implement WorldBridge based on GAE~\cite{Lu2026GAELA}, which provides a good geometry-appearance space for scene generation. 
The scene encoder and scene reader are initialized from LagerNVS~\cite{szymanowicz2026lagernvs} and tuned; the reference and text encoders are DepthAnything v3~\cite{depthanything3} and Qwen3-0.6B~\cite{yang2025qwen3} and kept frozen; the RGB and geometry decoder are kept frozen; the pipeline has 0.94B trainable parameters in total.
We train our model on a collection of video clips from RealEstate10K~\cite{zhou2018stereo}, DL3DV~\cite{ling2024dl3dv}, and SpatialVID~\cite{wang2025spatialvid}, which contains about 22k video clips at a $1280\times720$ resolution and is separate from the evaluation benchmarks.
The training data is labeled by ViPE~\cite{Huang2025ViPEVP} and
DepthAnything v3~\cite{depthanything3} for metric camera pose
and depth, and Qwen3.5-27B~\cite{qwen3.5} for caption.
The diffusion transformer is trained with a resolution of $672\times378$, while the global context supervision is at a resolution of $512\times288$.
The model is trained on 32 GPUs for 10 epochs, using the Adam optimizer with a batch size of 64, and a learning rate decaying from $5 \times 10^{-5}$ to $5 \times 10^{-6}$.

\subsection{Training Strategy}
\paragraph{Target-view sampling.} 
At each training step, we draw the target sequence length $N$ uniformly from $\{21,\ldots,41\}$. 
We use two sampling modes to expose the model to both wide camera motion and dense local trajectories.
With probability $0.8$, we partition the full clip into $N$ equal-duration bins and select one valid frame uniformly from each bin.
With probability $0.2$, we sample $N$ consecutive frames, using a randomly selected valid start and stride when a constant-stride sequence is available, and an irregular sequence of valid frames within the same duration range otherwise. 
In both modes, we order the target frames chronologically and use their sequence indices $0,\ldots, N-1$ for temporal RoPE.

\paragraph{Reference-view sampling.} 
Temporal references are selected from the sampled target sequence, while spatial references are drawn uniformly without replacement from the same clip's full pool of valid frames after excluding the sampled target frame indices. 
Visual overlap between spatial references and targets can vary across samples.
We sample up to 6 references during training, while it can infer with more references.
The temporal-only configurations use the first frame ($15\%$), both endpoints ($15\%$), or three to six target frames ($10\%$). 
The spatial-only configuration uses one to six spatial references ($35\%$). 
The mixed configurations use the first frame and one to five spatial references ($10\%$), or both endpoints and one to four spatial references ($10\%$). 
The remaining $5\%$ of the configuration distribution uses no image references. 
Within each configuration, any variable reference count is sampled uniformly over its specified range.
A separate $5\%$ joint dropout of reference, camera, and text conditioning provides the unconditional branch for classifier-free guidance.

\paragraph{Temporal position dropout.} 
With probability $0.5$, we disable temporal RoPE for temporal references and treat them as spatial references.
Target-to-target attention retains temporal RoPE. This exposes the model to the same reference images both with and without temporal positions, complementing training with spatial references, whose interactions always omit temporal RoPE.

\subsection{Inference Setting}
For the benchmark results reported in the main paper, we generate each target sequence in a single sampling run with 50 denoising steps, a classifier-free guidance scale of 2, a seed of 42, and an empty text prompt.
Let $N$ be the number of requested target views. When $N \leq 41$, we insert $41-N$ auxiliary cameras into the target trajectory and generate all 41 views jointly.
We evaluate the outputs at the original target cameras.
When $N>41$ (e.g., $N=56$ for WorldBridge-Long and WorldBridge-Agent (On), and $N=50$ for WorldBridge-Agent (Off)), we extend the temporal RoPE indices from $0,\ldots,40$ to $0,\ldots, N-1$ and generate all $N$ views.
For longer trajectories, WorldBridge also supports segment-wise generation. The final $K$ generated frames of each segment serve as temporal context for generating the next segment.

\subsection{Latent-reference Consistency Configuration}

For the implementation of Latent-reference Consistency, we estimate each reference's metric depth once with MoGe-3~\cite{Kong2026MoGe3FM} using the supplied camera field of view, and a single refinement step in each denoising step.
Using the supplied cameras, we forward-project the reference RGB into each target view and retain valid projected pixels. 
The gradient direction is RMS-normalized, with a relative step size of 0.01. 
For view $i$, let $d_i\in[0,1]$ be its distance to the nearest reference (if the reference is a spatial reference, it should be the target view that is nearest to the spatial reference), normalized by the maximum over views. 
At noise level $t$, the fresh-noise mixing coefficient is $\lambda_i(t)=0.1t+(0.5-0.1t)d_i$, capped at 0.5. 
Thus, the transition mixes estimated and fresh noise with amplitudes $\sqrt{1-\lambda_i}$ and $\sqrt{\lambda_i}$, respectively.

\paragraph{Inference time.} We evaluate the inference efficiency of a 41-frame sequence generation on a single H200. Under 50 denoising steps, seed 42, and CFG 2, using LRC costs 18.3\% more time compared with inference without LRC.

\section{WorldBridge Benchmark Details}
\label{sec:benchmark}

Our WorldBridge benchmark suite comprises 100 short real-world video clips, 20 long real-world video clips, and 20 agentic-authored worlds.
Each authored world contributes two paired evaluation cases, On and Off, for a total of 40 cases.
We will release our benchmark as well as our model upon acceptance.

\paragraph{Real-video curation and annotation.} 
The curation protocol uses two stages. 
First, Qwen3.5-27B~\cite{qwen3.5} separately reviews a chronological sequence of sampled RGB frames and the candidate reference images.
The first review checks shot continuity, distracting overlays, visual quality, and independently moving objects. The second checks whether the references provide complementary observations of a coherent scene. 
RoMa v2~\cite{edstedt2025romav2} measures matching support for the single Short pair or all 15 Long pairs. 
The protocol advances candidates to reconstruction when both visual checks and the matching criteria pass.

In the second stage, ViPE~\cite{Huang2025ViPEVP} estimates per-frame camera pose and intrinsics along the retained clip. DA3 Nested~\cite{depthanything3} uses the estimated cameras and intrinsics to predict the depth maps. 
One positive scale per clip rescales the ViPE~\cite{Huang2025ViPEVP} translations to the metric scale of the predicted depth. 
We check frame correspondence, valid depth coverage, abrupt pose changes, scale consistency across windows, and agreement between overlapping depth estimates. 
We review RGB, depth, and trajectory overlays and hold clips with unresolved issues. 
Captions describe the scene and camera motion.

\subsection{WorldBridge-Short}

The 100 clips come from RealEstate10K~\cite{zhou2018stereo} (29 clips), DL3DV~\cite{ling2024dl3dv} (37 clips), and SpatialVID~\cite{wang2025spatialvid} (34 clips). 
All clips' durations are 5--15 seconds, and the first and last frames provide two references. 
The retained clips include 36 indoor, 20 outdoor, 23 indoor-to-outdoor, and 21 outdoor-to-indoor paths.

We back-project candidate target pixels with the estimated target depth, transform them to each reference camera, and check image bounds and agreement with the reference depth. We take the union of the two reference support masks. Eligible intermediate frames have at least 50\% estimated reference support. Viewpoint diversity and near-duplicate checks reduce repeated views, and we retain at most six targets per clip. This yields 549 target views across 100 clips, with two to six views per clip.

\begin{figure*}[t]
    \centering
    \includegraphics[width=1.\linewidth]{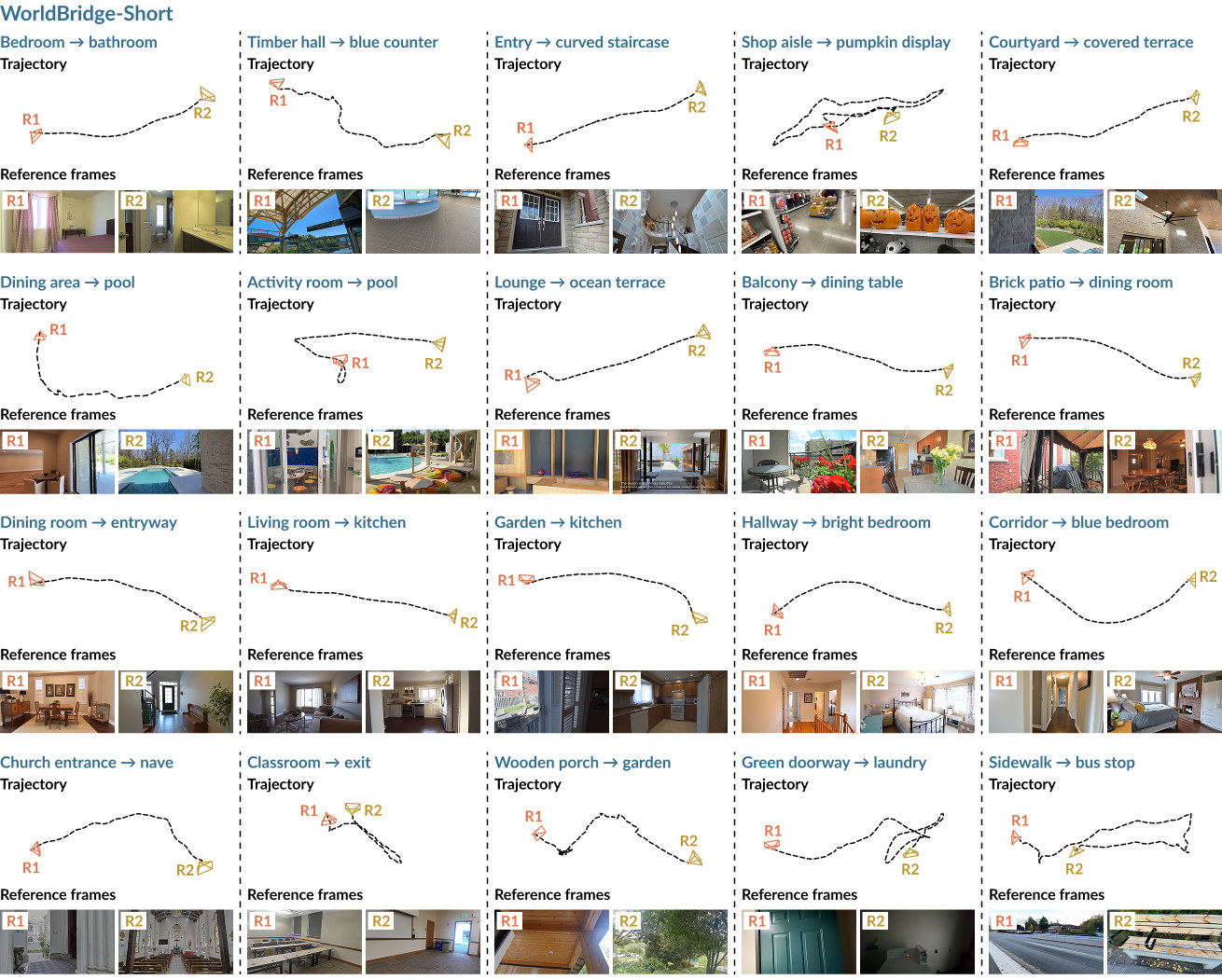}
    \caption{More visualizations of WorldBridge-Short samples.}
    \label{fig:supp_wb_short}
\end{figure*}

\subsection{WorldBridge-Long}

The 20 retained long clips comprise 10 indoor and 10 outdoor scenes.
The evaluated spans range from 20 to 60.0 seconds. Six reference frames follow one continuous trajectory and cover complementary regions. 
We place ten evenly spaced interior timestamps in each of the five reference intervals and use the nearest retained source frames, excluding the references. This gives 50 evaluation views per clip.
For labeling the long clips, we sample native frames at approximately 15 fps while preserving the exact six reference frames and their timestamps. We estimate dense depth in overlapping windows and apply a single translation scale across the clip.

\begin{figure*}[t]
    \centering
    \includegraphics[width=1.\linewidth]{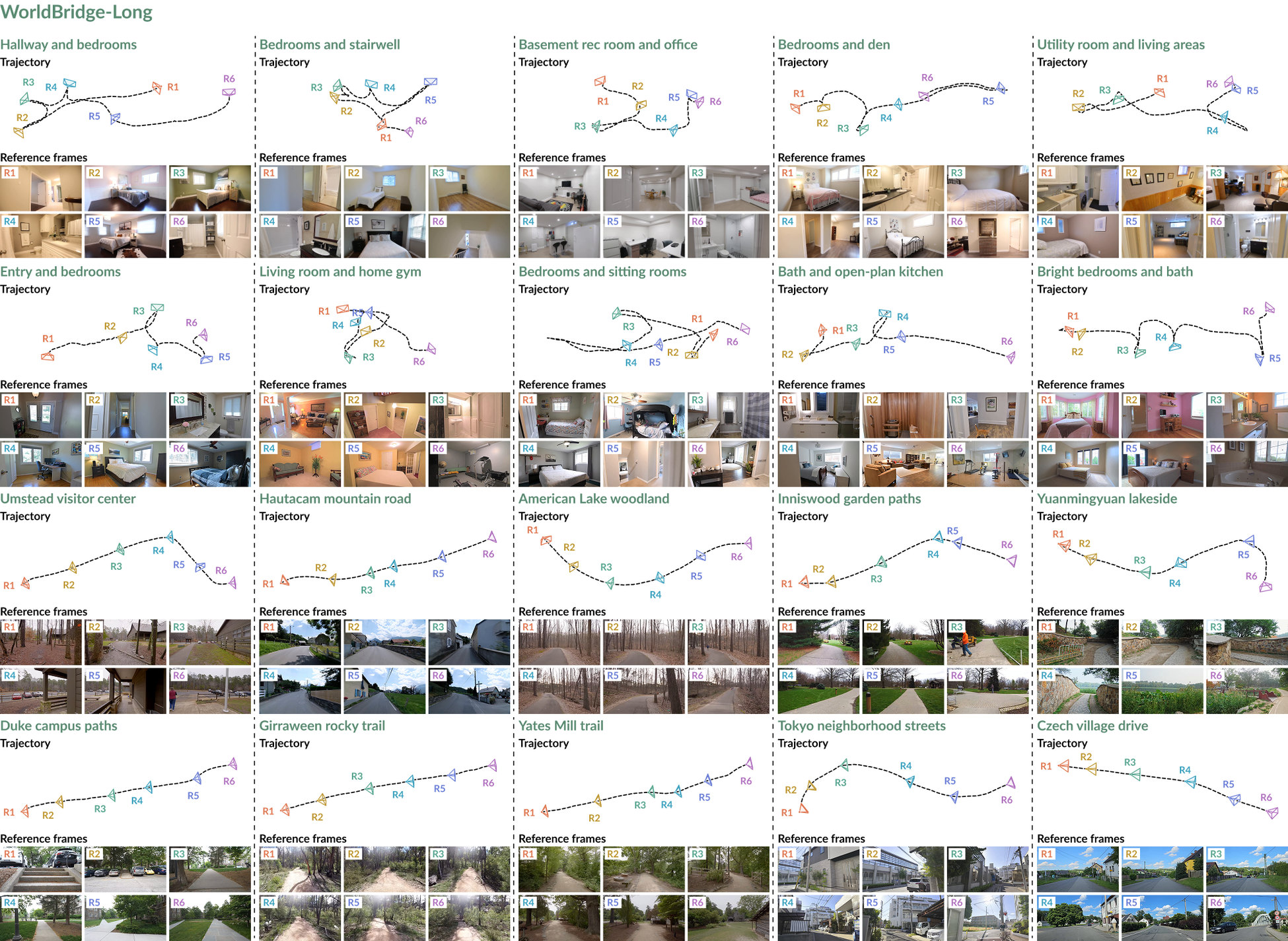}
    \caption{More visualizations of WorldBridge-Long samples.}
    \label{fig:supp_wb_long}
\end{figure*}

\subsection{WorldBridge-Agent}

\paragraph{World construction.} The 20 agent-authored worlds comprise 10 indoor and 10 outdoor environments. 
Indoor examples include offices, libraries, exhibition spaces, and residences. 
Outdoor examples include streets, markets, gardens, and coastal trails. 
We use agent-assisted geometric blockouts to specify the layout, traversable regions, and camera paths. 
We render structural guides at the reference cameras and use them to generate images with a shared scene description and appearance cues. 
We compare the generated views against the guides and other references, checking openings, major structures, object identity, and plausible sightlines. 
Off-path camera placement also checks clearance, reachability, and visibility in the blockout.
In our construction, the agent is GPT-6 Astra~\cite{openai2026astra}, and the image generation model is GPT-Image~\cite{openai2026images25}.

\begin{figure*}[t]
    \centering
    \includegraphics[width=1.\linewidth]{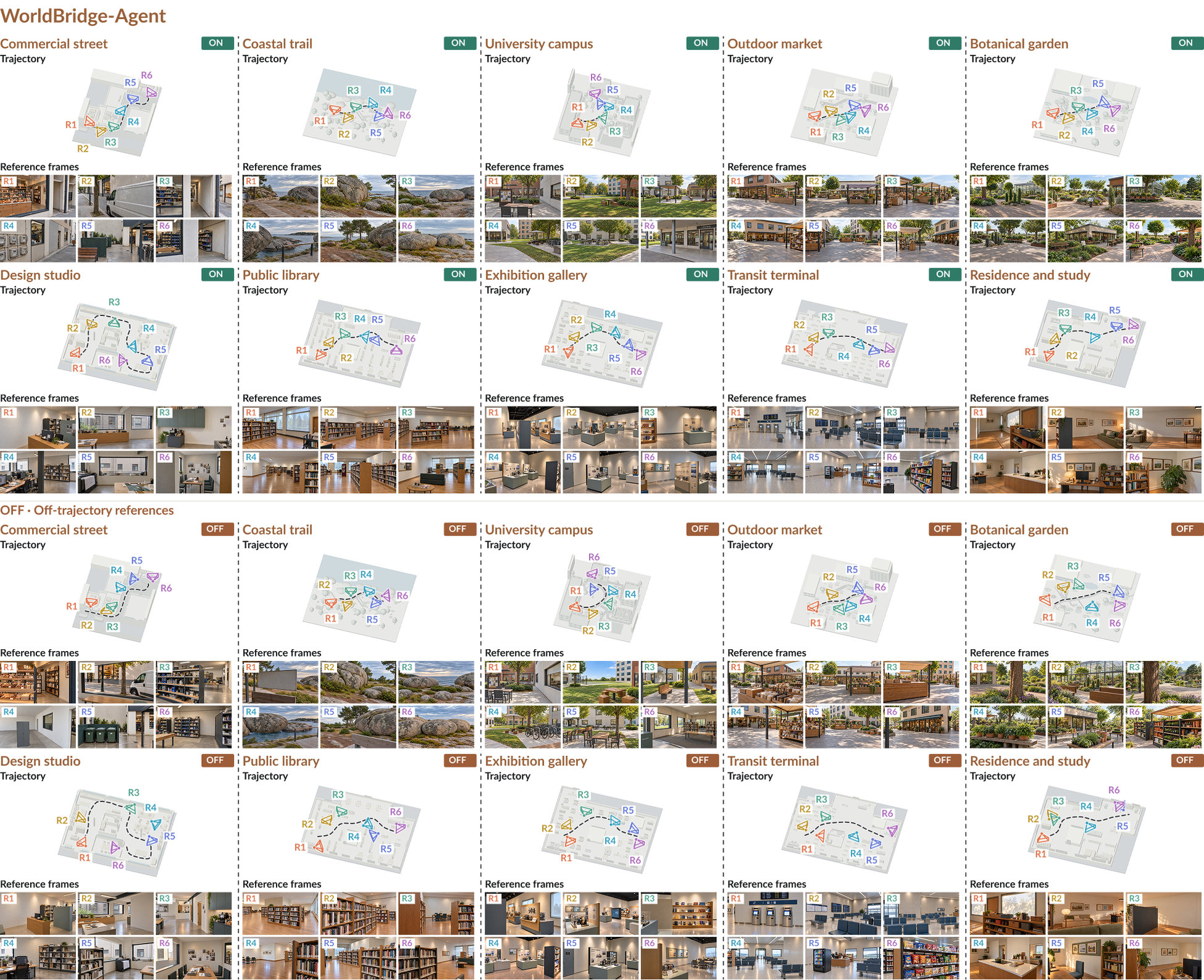}
    \caption{More visualizations of WorldBridge-Agent samples.}
    \label{fig:supp_wb_agent}
\end{figure*}

\paragraph{Paired On and Off conditions.} 
Both conditions share the same authored world and target trajectory. 
On-trajectory provides six reference images at poses on the trajectory, with known temporal positions. 
Off-trajectory provides six separately generated references at poses outside the target path, with spatial camera parameters and no assigned timestamps. 
We select 50 common target cameras by taking the nearest authored poses to ten evenly spaced interior times in each interval between successive On references. 
The nominal trajectories span 50--58 seconds. 
The authored camera parameters specify the intended views, and the structural guides constrain reference generation.

\subsection{Benchmark Visualizations}

We provide more visualization examples of our WorldBridge benchmark suite in \cref{fig:supp_wb_short,fig:supp_wb_long,fig:supp_wb_agent} to show the diversity of our benchmark.

\paragraph{Reference overlap.} 
We report RoMa v2 matching coverage for the reference inputs.
For each direction of a reference pair, we count pixels with in-bounds forward and backward matches, confidence at least 0.5 in both directions, and a round-trip error of at most two pixels on a $512\times288$ grid. 
We divide by the size of the full source grid and average both directions. 
There is 1 pair per short clip and 15 pairs per six-reference case.
The quantitative results are shown below.

\begingroup
\setlength{\intextsep}{5pt plus 1pt minus 1pt}
\begin{table}[!ht]
\centering
\captionsetup{position=top,skip=3pt}
\caption{The matching coverage distribution of each subset of our WorldBridge benchmark suite.}
\footnotesize
\setlength{\tabcolsep}{3pt}
\begin{tabular}{@{}lrrrrr@{}}
\toprule
Subset & Pairs & Mean & $[0,1]\%$ & $(1,5]\%$ & $>5\%$ \\
\midrule
Short     & 100 & $0.368\%$ & 93 (93.0\%) & 4 (4.0\%) & 3 (3.0\%) \\
Long      & 300 & $0.837\%$ & 258 (86.0\%) & 24 (8.0\%) & 18 (6.0\%) \\
Agent On  & 300 & $0.017\%$ & 300 (100.0\%) & 0 (0.0\%) & 0 (0.0\%) \\
Agent Off & 300 & $0.085\%$ & 296 (98.7\%) & 3 (1.0\%) & 1 (0.3\%) \\
\bottomrule
\end{tabular}
\label{table:bench_match}
\end{table}
\endgroup

\section{Evaluation Protocol}
\label{sec:eval_protocol}

\subsection{Baseline Adaptation}We use the officially released checkpoints and inference pipelines for all baselines. We adapt the input format to each method, convert the supplied poses to its camera convention, and rescale the intrinsics to its input resolution. The following protocols describe the details.

\paragraph{SEVA~\cite{zhou2025stable}.}
SEVA is implemented with 50 denoising steps and a classifier-free guidance (CFG) scale of 2. 
We use its official \texttt{img2vid} when the reference cameras lie on the trajectory, placing reference images in trajectory order. 
For off-trajectory references, we use \texttt{img2trajvid}, with the posed references supplied before the ordered target cameras.
For 50-view evaluations, we use SEVA's official procedural two-pass sampling, generating sparse anchors first and each chunk later.
\paragraph{GAE~\cite{Lu2026GAELA}.}
We use the released D64-1B model, its geometry autoencoder, and the Euler flow sampler with 50 steps and CFG 2. 
In the evaluation of more-than-one-reference-images, we extend the released single-reference configuration by encoding multiple reference images jointly in one reference-only pass. 
We then sample the 50 target views and 31 auxiliary views jointly in one call, with the clean reference latents provided as additional conditions at their supplied camera poses.
For off-trajectory references, we retain the camera poses of all references and assign them a shared temporal RoPE position at the midpoint of the normalized trajectory, since their timestamps are unavailable.

\paragraph{Gen3R~\cite{Huang2026Gen3R3S}.}
We use the released image-to-world pipeline, including its Pl\"ucker ray conditioning, with 49 frame slots, 50 denoising steps, and CFG 5.
As Gen3R does not support more than two sparse references as input, we retain the supported one- or two-image conditioning and place two references in the endpoint slots to generate different chunks.
For multiple-reference-guided generation, each call uses two reference images and auxiliary camera views.
For each target segment in the off-trajectory setting, we select the
two references with the shortest distance to the target
cameras in that segment, assigning these references to the first and last frame slots for the generation of one interval.

\paragraph{GLD~\cite{jang2026gld}.}
We use the released DA3-based latent diffusion cascade with Pl\"ucker rays and ProPE camera conditioning. 
Each call has eight view slots, with reference latents filling the conditioned slots and target latents sampled in the remaining slots. 
We use 50 denoising steps per stage and CFG 1.5.
With six reference images in WorldBridge-Long and WorldBridge-Agent, each call generates two adjacent target views. We repeat this for all 25 pairs using the same six original references.
\paragraph{AnyRecon~\cite{chen2026anyrecon}.}
We use the released point-cloud-guided video refinement pipeline after DMD~\cite{DMD}, with four denoising steps and CFG 1.
We reconstruct the conditioning point cloud with Pi3~\cite{wang2025pi} using the reference RGB images,
align the estimated geometry to the provided reference cameras and render it with visibility masks at the target cameras.
The original reference images are prepended to the rendered target sequence for video refinement. 
For WorldBridge-Long and WorldBridge-Agent, it
divides the rendered views into consecutive chunks of 35. 
For each chunk, we follow its official setting, which uses all reference images, and insert the last two frames generated from the previous chunk as auxiliary reference images to provide context across chunk boundaries.
\paragraph{3DRAE~\cite{wei20263drae}.}
We use the released scene-latent generator with 50 Euler steps and CFG 2. 
The appearance condition consists of the reference RGBs at their supplied camera poses. 
For WorldBridge-Long and WorldBridge-Agent, the trajectory condition combines the six reference cameras with ten camera-only anchors sampled along the target path. 
We order on-trajectory references by their timestamps and associate off-trajectory references with nearby anchors using a camera-pose distance. 
After the generation process of the scene latent, it queries its decoder at all 50 target cameras to obtain RGB images and point maps.

\subsection{Evaluation Settings}
We use the same target cameras and available reference images for all methods within each sample.
Auxiliary views used during generation are excluded from scoring. 
For the public novel-view-synthesis benchmarks, we score seven target views with a single reference and six target views with two endpoint references. The evaluation of metrics is conducted at a resolution of $512\times288$.
We score the selected intermediate views on WorldBridge-Short and 50 non-reference target views per scene on WorldBridge-Long and WorldBridge-Agent.

\subsection{RGB Fidelity and Native Geometry Metrics}
\paragraph{RGB fidelity.} 
We compare generated and ground-truth RGB at the same target view using PSNR, SSIM, and LPIPS. We resize both images to a common resolution of $512\times288$ for metrics computation. 

\paragraph{Native geometry.} 
On WorldBridge-Long, we compare each method's native geometry output with the pseudo ground-truth depth from ViPE~\cite{Huang2025ViPEVP} and DepthAnything v3~\cite{depthanything3}. 
We filter low-confidence pixels below the first quartile of positive confidence, and exclude depth discontinuities above a 10\% relative jump and the image boundary. 
For each method's output geometry of a scene, we fit a single scale $s_{\mathrm{d}}$ as the median of the source-to-predicted depth ratios pooled over all valid target pixels. 
We compute AbsRel as the mean absolute relative error $|s_{\mathrm{d}}\hat d-d|/d$ on valid predicted pixels within the source mask. 
For $\delta_1$, we report the fraction of the full source mask satisfying $\max(s_{\mathrm{d}}\hat d/d,d/(s_{\mathrm{d}}\hat d))<1.25$, with invalid predictions counted as failures.

\subsection{Camera Control and Consistency Metrics}
\paragraph{Camera control.} 
For WorldBridge-Long and WorldBridge-Agent, we jointly reconstruct the 50 generated target views with VGGT-$\Omega$~\cite{wang2026vggtomega} to estimate their poses and depths. 
We resize images to $560\times304$ to input to the reconstruction model, and fit one positive-scale similarity transform to the estimated and ground truth camera centers. 
Using the resulting aligned trajectory, we compute the translational RMSE of the absolute trajectory error (ATE), and the rotational and translational RMSEs of relative pose error (RPE) over adjacent target views with the pseudo ground-truth. 
Rotation is measured in degrees and translations in the units of the labeled trajectory. 
For WorldBridge-Agent, the supplied cameras are the agentic nominal cameras, which are designed as metric scale.

\paragraph{Cross-view consistency.} 
To measure consistency under the requested camera motion, we scale the reconstructed depths from VGGT-$\Omega$~\cite{wang2026vggtomega} by the fitted scale $s$ and back-project using the provided intrinsics $K_i$ and camera-to-world pose $(R_i,C_i)$ by
\begin{equation}
    X_i(u)=R_i\big(s\hat D_i(u)K_i^{-1}\bar u\big)+C_i,
\end{equation}
where $\bar u$ is the homogeneous pixel coordinate. 
Then, we project each view into its adjacent target in both directions. 
Geometric reprojection uses a nearest-depth buffer and averages $|z_{i\to j}-z_j|/\max(z_{i\to j},z_j)$ where both depths are valid. 
Feature reprojection forward-splats dense DINO ViT-S/16~\cite{Caron_2021_ICCV_dino} features upsampled with FeatUp~\cite{fu2024featup}, retaining contributions within 5\% of the nearest projected depth. 
We average the cosine distance to the destination features over valid projected pixels. 
We require at least 64 valid pixels per direction and exclude near-constant images. 
Geometric errors are averaged over supported directions, and feature errors over pairs with both directions supported.

\subsection{Reference Fidelity and Quality Metrics}

\paragraph{Reference fidelity.}
Each reference is jointly reconstructed with eight nearest generated target views in a separate nine-image VGGT-$\Omega$~\cite{wang2026vggtomega} pass. 
For each group, we fit its scale with the provided cameras and use the provided cameras to project the reference RGB into those target views with the depth-aware splatting rule above. 
Given at least 64 valid projected pixels, we set both images to zero outside the projection mask, compute spatial LPIPS, and average its distance map within the mask. 
We average supported directions within each reference and then average the reference scores within the scene.

\begin{figure*}[t]
    \centering
    \includegraphics[width=1.\linewidth]{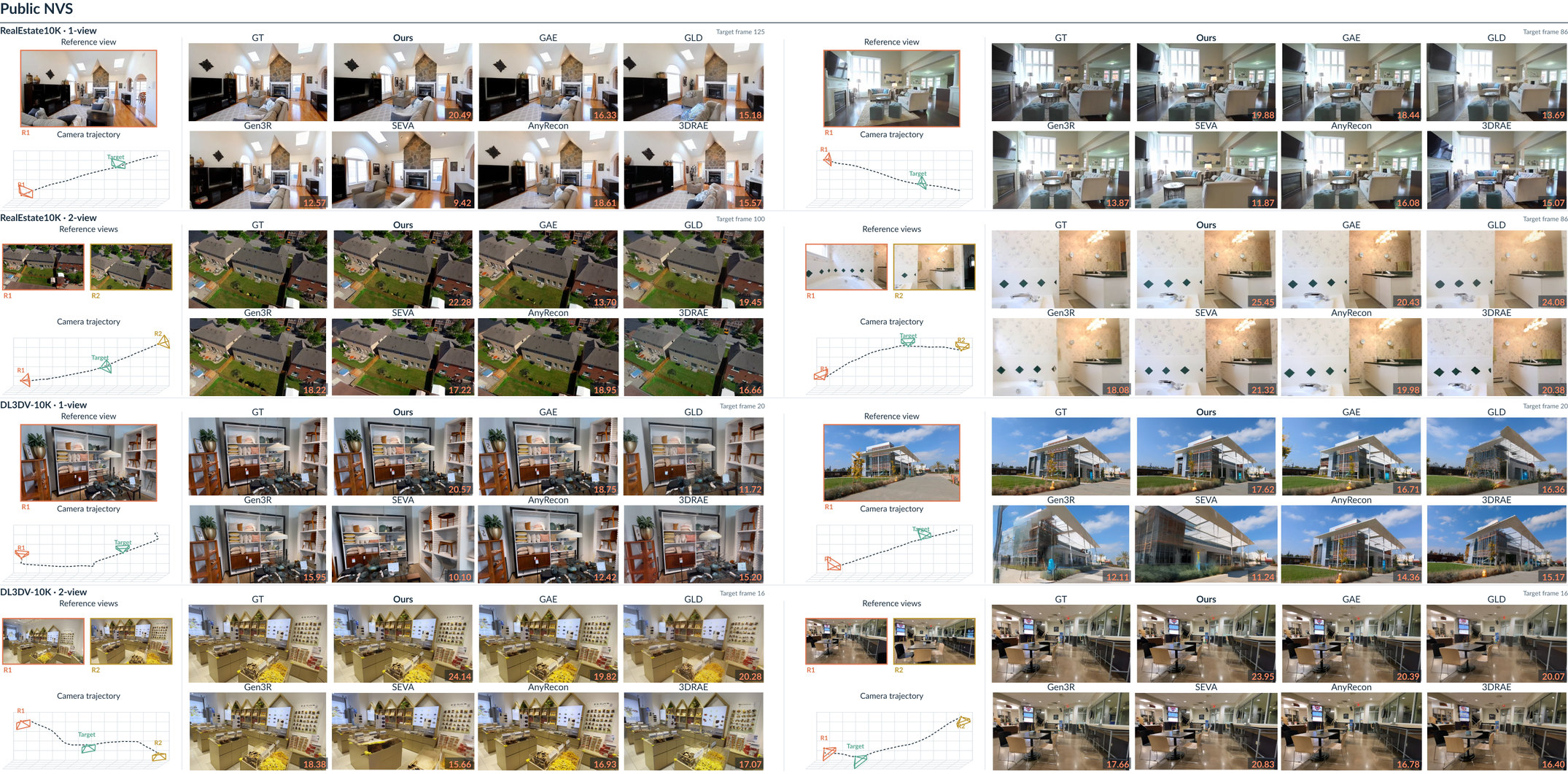}
    \caption{More qualitative results of the public NVS benchmarks~\cite{zhou2018stereo,ling2024dl3dv}.}
    \label{fig:supp_qualitative_public_nvs}
\end{figure*}

\begin{figure*}[t]
    \centering
    \includegraphics[width=1.\linewidth]{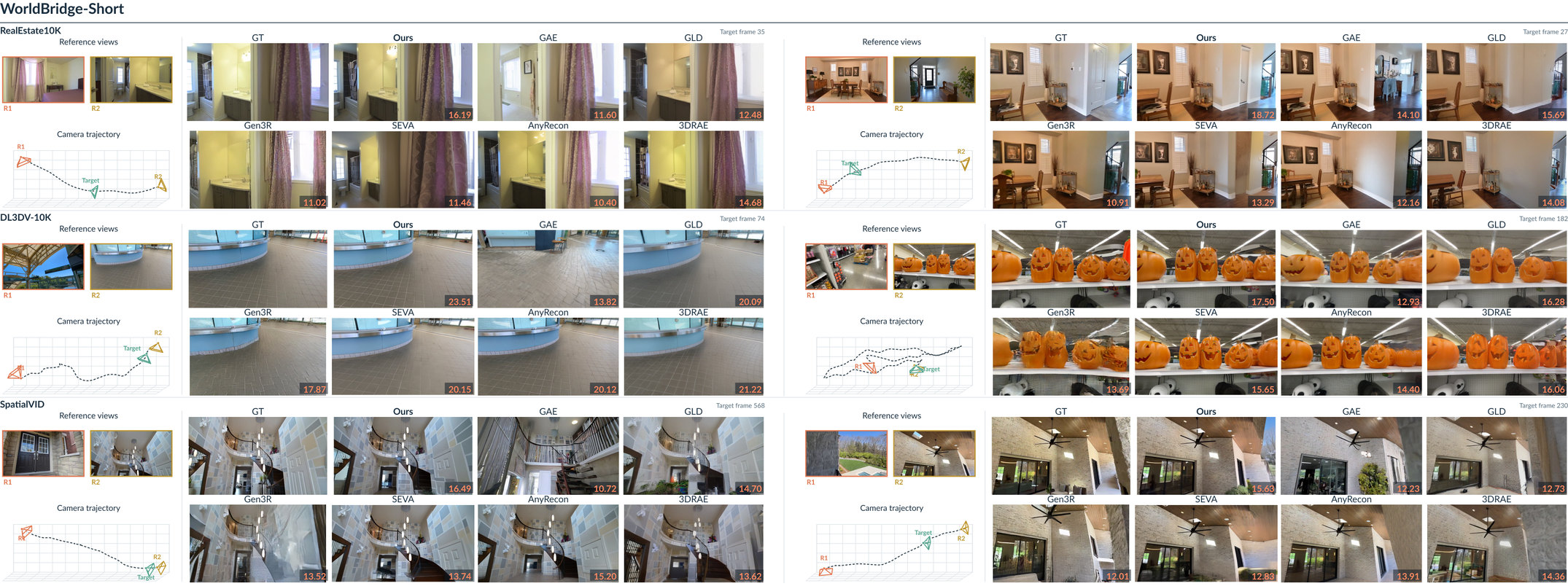}
    \caption{More qualitative results of WorldBridge-Short.}
    \label{fig:supp_qualitative_wb_short}
\end{figure*}

\paragraph{Aesthetic quality.} 
We score each generated target frame with a linear aesthetic predictor applied to the normalized CLIP ViT-L/14~\cite{Clip} image embedding. Inputs are resized to $512\times288$ and then processed by CLIP's $224\times224$ center-crop pipeline.

\section{Additional Results}
\label{sec:additional_results}

\begin{figure*}[t]
    \centering
    \includegraphics[width=1.\linewidth]{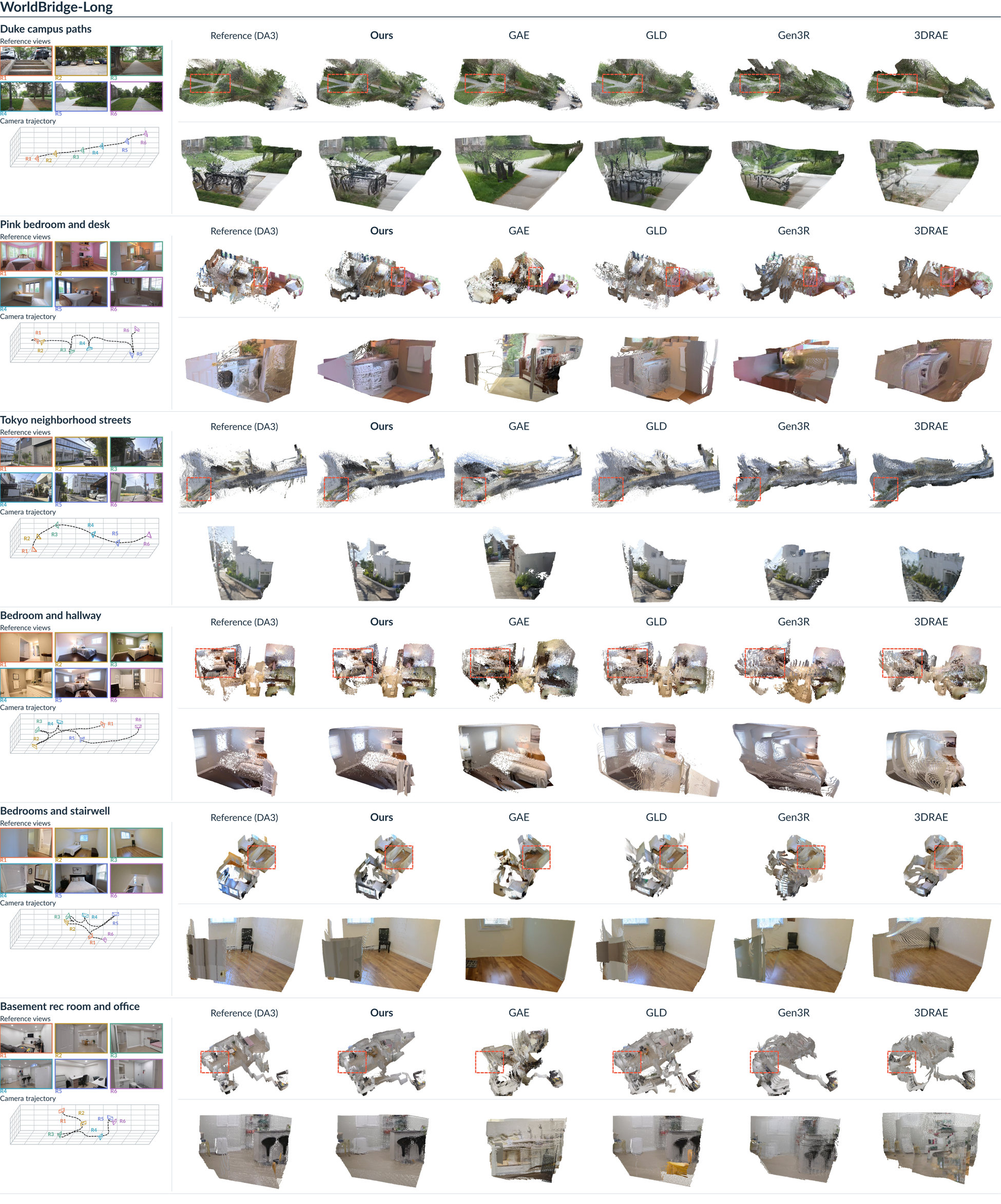}
    \caption{More qualitative results of WorldBridge-Long.}
    \label{fig:supp_qualitative_wb_long}
\end{figure*}

\begin{figure*}[t]
    \centering
    \includegraphics[width=1.\linewidth]{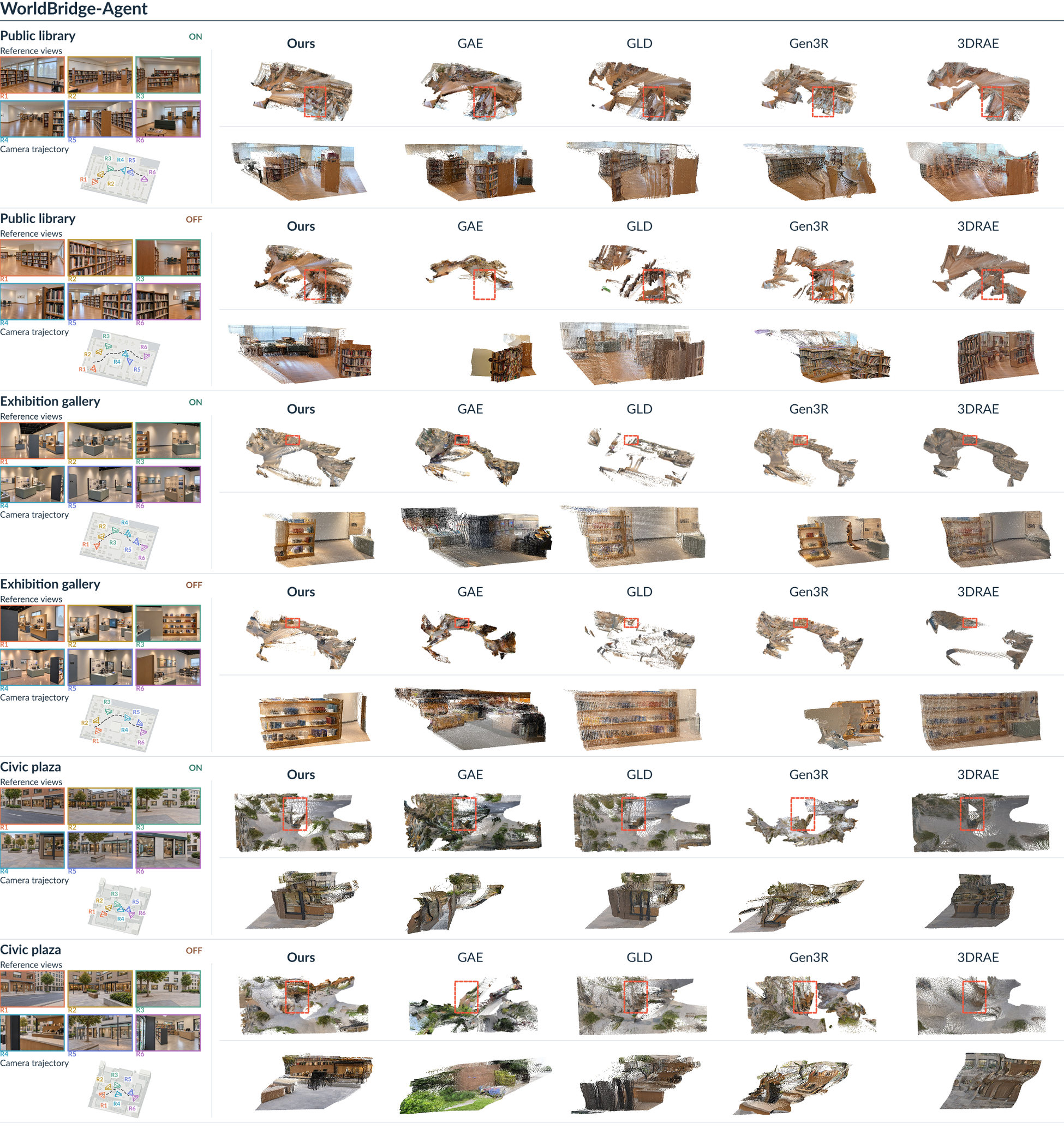}
    \caption{More qualitative results of WorldBridge-Agent.}
    \label{fig:supp_qualitative_wb_agent}
\end{figure*}

\subsection{Qualitative Comparisons}
To verify the effectiveness of our method, we provide more visualization results of different benchmarks.
On RealEstate10K~\cite{zhou2018stereo} and DL3DV~\cite{ling2024dl3dv} as shown in Fig.~\ref{fig:supp_qualitative_public_nvs}, the input includes one or two references, usually with obvious visual overlap. We provide the visual results as well as the PSNR metric to show our superiority.
On our proposed WorldBridge-Short, the two references always share little or even no visual overlap (Fig.~\ref{fig:supp_qualitative_wb_short}), which is much more challenging. 
Our method provides results with more reasonable and coherent content to bridge the scattered reference views, with obviously higher PSNR.

With six references, the point clouds reveal whether structures seen at different positions along a trajectory fit together. 
On WorldBridge-Long (Fig.~\ref{fig:supp_qualitative_wb_long}), our model produces much more consistent 3D scenes with better fine-grained geometry details.
For instance, in the first row, the campus path and the bicycles beside it are both visible in our reconstruction and in the reference DA3 reconstruction of the captured views. Several baselines recover the path but omit or distort the bicycles' geometry details. 
As shown in Fig.~\ref{fig:supp_qualitative_wb_agent}, WorldBridge-Agent assesses a same target path with references either on or off the trajectory. Our method can also provide a more coherent 3D scene. Taking the public library samples, the bookshelves stand on a continuous floor, and the enlarged views show the space between the shelves, which shows much better structure and details compared with the baselines, as well as following the reference's guidance.

\begin{figure*}[!t]
    \centering
    \includegraphics[width=1.\linewidth]{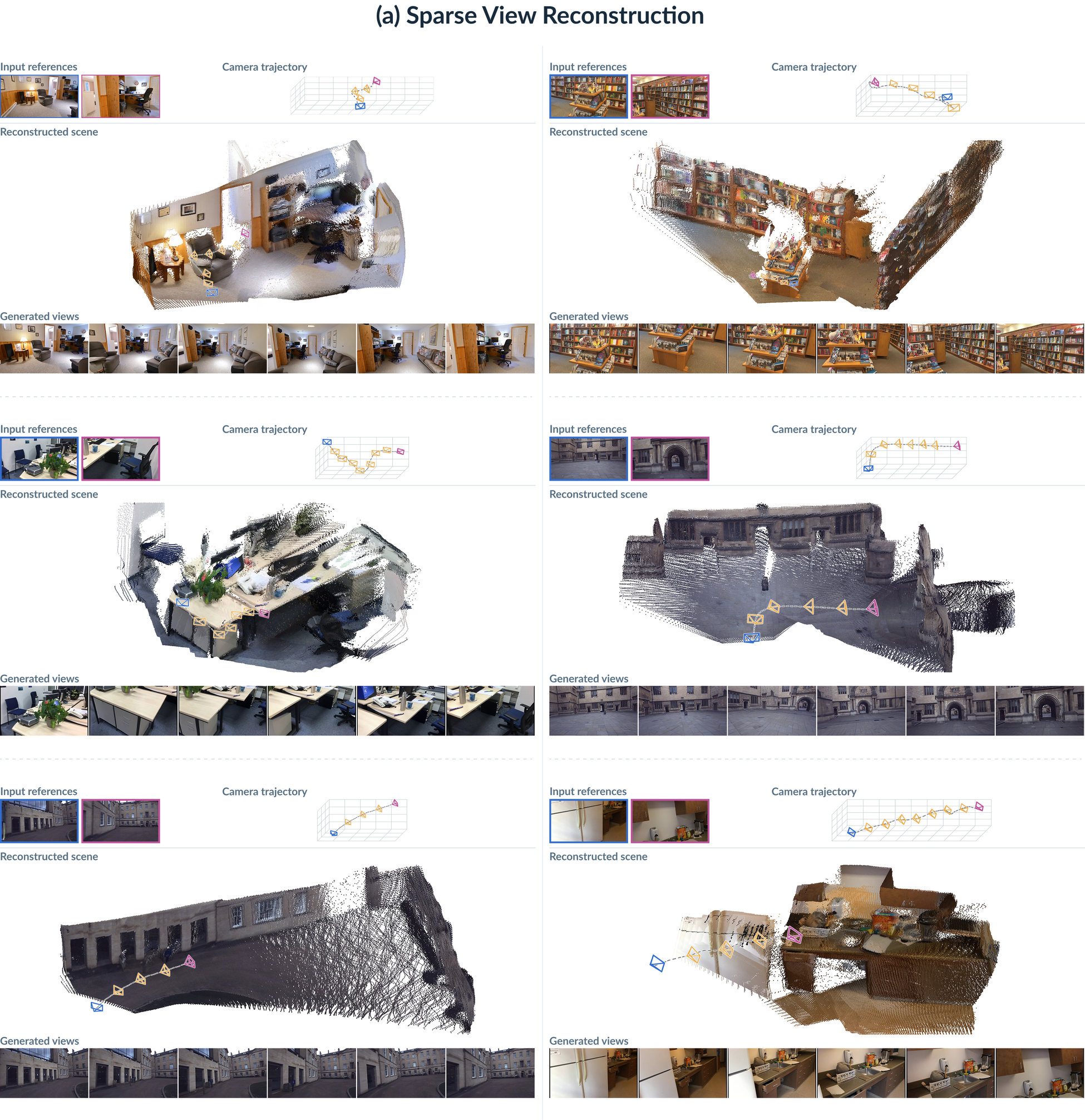}
    \caption{More visualizations of sparse reconstruction by WorldBridge.}
    \label{fig:supp_more_sparse_recon}
\end{figure*}

\subsection{Extended Applications}
\begin{figure*}[!t]
    \centering
    \includegraphics[width=1.\linewidth]{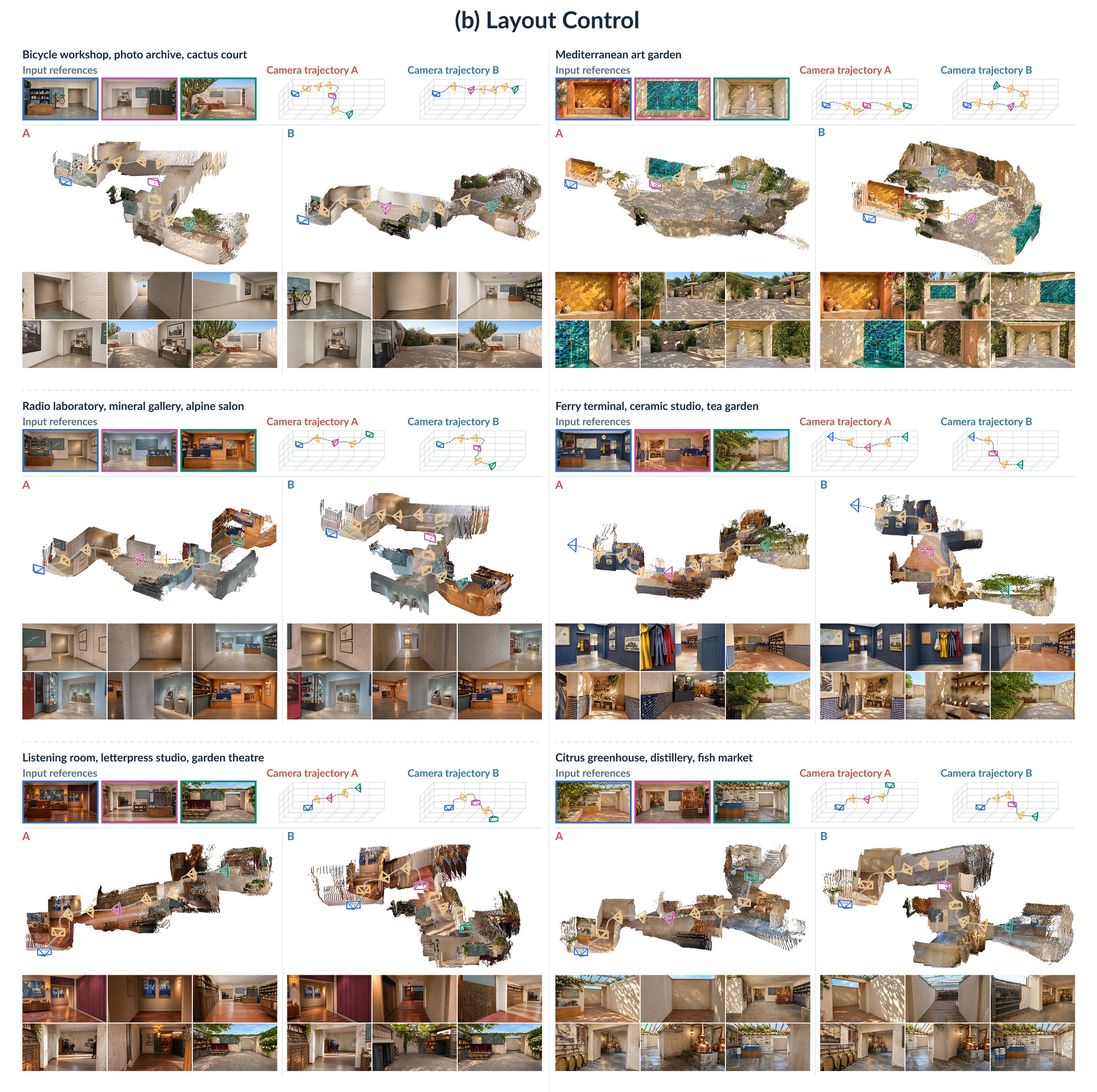}
    \caption{More visualizations of layout control by WorldBridge.}
    \label{fig:supp_more_layout_control}
\end{figure*}

\begin{figure*}[!t]
    \centering
    \includegraphics[width=1.\linewidth]{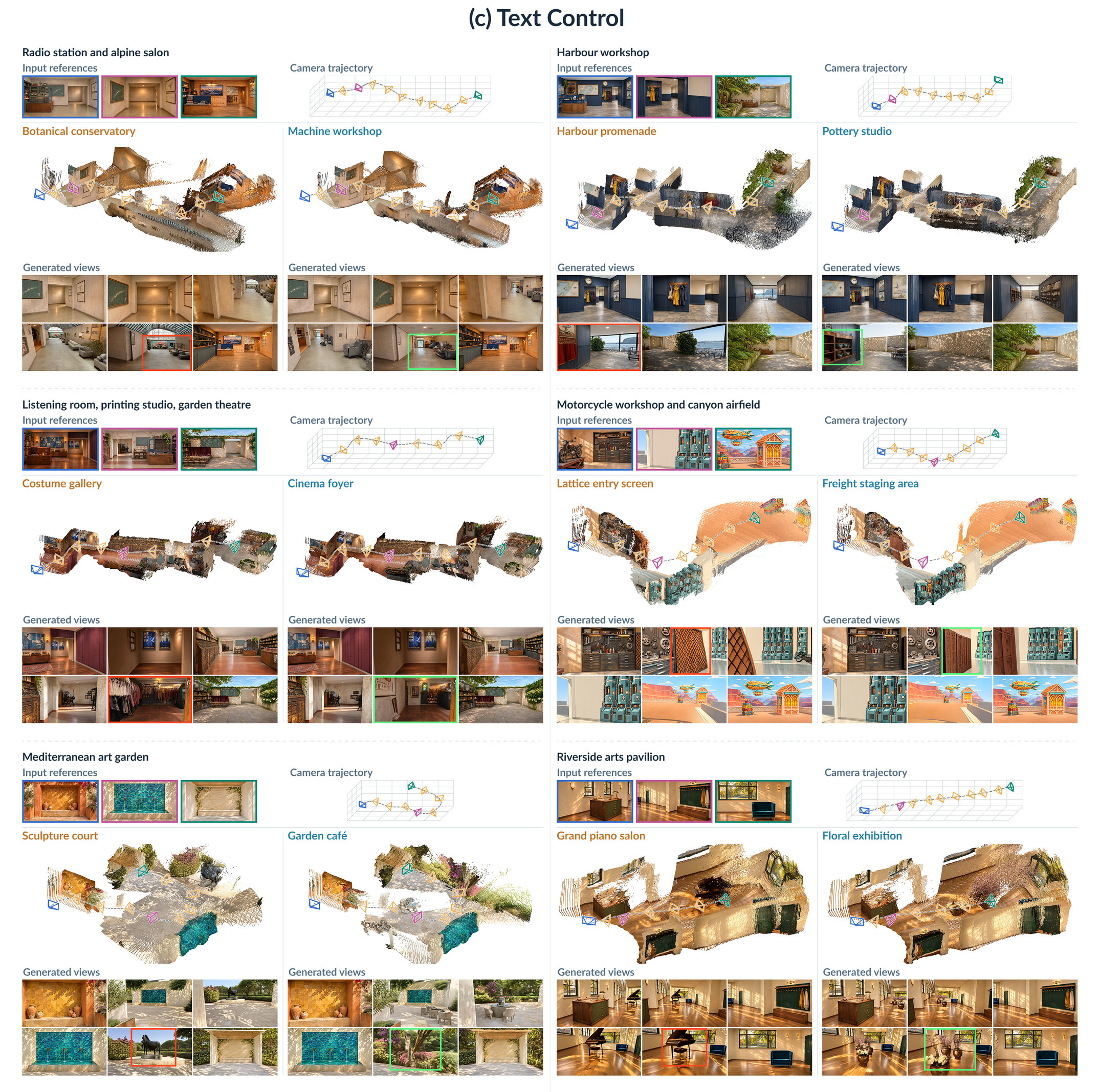}
    \caption{More visualizations of text control by WorldBridge.}
    \label{fig:supp_more_text_control}
\end{figure*}

To verify different application capabilities mentioned in Section 4.5 in the main paper, we provide more visual examples about sparse reconstruction (Fig.~\ref{fig:supp_more_sparse_recon}), layout control (Fig.~\ref{fig:supp_more_layout_control}), text control (Fig.~\ref{fig:supp_more_text_control}), and different visual styles bridging (shown in our project page).

\paragraph{Sparse-view scene reconstruction.} 
Given posed photographs and a target trajectory, WorldBridge generates RGB views and corresponding point maps. In the bookshop, one reference looks across a freestanding display and the other looks along the wall shelves. The intermediate views and the assembled point cloud place the display within the aisle, so the viewer can move between the two observations. More results can be seen in Fig.~\ref{fig:supp_more_sparse_recon}.

\paragraph{Layout-control through camera poses.} 
Users can specify where reference content should appear in a new world by changing the relative poses assigned to the same reference images. For instance, in the workshop, photo archive, and cactus court example, the two camera configurations produce different turns and connecting spaces while retaining all three locations. More details and examples can be seen in Fig.~\ref{fig:supp_more_layout_control}.

\begin{samepage}
\paragraph{Text-control in unseen regions.}
With the reference images, poses, and the target trajectory fixed, a prompt can specify what occupies an unobserved region between different observed references. 
For example, in the sample of riverside pavilion, the same stretch of the route opens onto a grand piano or a floral arrangement. The window and teal chair from the reference remain recognizable in both versions. 
More details and examples can be seen in Fig.~\ref{fig:supp_more_text_control}.
\par
\end{samepage}

\paragraph{Bridging real and imaginary worlds.} 
References can also depict different visual styles. 
We provide more visualization results on our project page.
The offline viewer includes 8 paired on- and off-trajectory imaginary world exploration videos, and real-world sequences from ScanNet v2~\cite{dai2017scannet} and Oxford Spires~\cite{tao2025spires}.
Across the shown videos, WorldBridge preserves much better details from the input references and maintains a much more consistent scene layout as the camera moves.

\subsection{More evaluations}

We evaluate the native target-view depth of the real-world examples in the offline webpage. The gallery contains 5 ScanNet v2 scenes~\cite{dai2017scannet} and 8 Oxford Spires scenes~\cite{tao2025spires}, with 50 target views per scene.
We compare the native generated depth accuracy on a common $512\times288$ grid with identical target views and sensor masks for all methods. For each method and scene, we fit one positive scale as the median ratio of sensor depth to predicted depth over valid pixels. We average the resulting AbsRel and $\delta_1$ over views within each scene, then weight scenes equally within each dataset. Tab.~\ref{tab:supp_real_native_depth} reports the quantitative results.

\begin{table}[!htb]
    \centering
    \footnotesize
    \setlength{\tabcolsep}{1.2pt}
    \caption{Scale-aligned native depth accuracy on the real-world examples from ScanNet v2~\cite{dai2017scannet} and Oxford Spires~\cite{tao2025spires}.}
    \label{tab:supp_real_native_depth}
    \begin{tabular*}{\columnwidth}{@{\extracolsep{\fill}}lcccccc@{}}
        \toprule
        & \multicolumn{2}{c}{ScanNet v2} & \multicolumn{2}{c}{Oxford Spires} & \multicolumn{2}{c}{Average} \\
        \cmidrule(lr){2-3}\cmidrule(lr){4-5}\cmidrule(l){6-7}
        Method & AbsRel $\downarrow$ & $\delta_1$ $\uparrow$ & AbsRel $\downarrow$ & $\delta_1$ $\uparrow$ & AbsRel $\downarrow$ & $\delta_1$ $\uparrow$ \\
        \midrule
        3DRAE~\cite{wei20263drae}         & 0.269 & 58.1 & \underline{0.449} & \underline{51.3} & \underline{0.359} & \underline{54.7} \\
        Gen3R~\cite{Huang2026Gen3R3S} & 0.289 & 54.9 & 0.530 & 37.0 & 0.409 & 45.9 \\
        GAE~\cite{Lu2026GAELA}           & \underline{0.265} & \underline{66.4} & 0.765 & 39.0 & 0.515 & 52.7 \\
        GLD~\cite{jang2026gld}           & 0.343 & 46.0 & 0.590 & 40.0 & 0.466 & 43.0 \\
        \midrule
        \textbf{Ours} & \textbf{0.168} & \textbf{79.1} & \textbf{0.378} & \textbf{65.9} & \textbf{0.273} & \textbf{72.5} \\
        \bottomrule
    \end{tabular*}
\end{table}

What's more, we also assess the cross-frame native geometry consistency on ScanNet v2~\cite{dai2017scannet}, Oxford Spires~\cite{tao2025spires}, and our proposed WorldBridge-Long, as these benchmarks have ground-truth or pseudo ground-truth geometry, which can be utilized for global alignment of all the approaches. Since WorldBridge-Agent does not have such geometry ground-truth, it is unfair to align each method's output geometry without a shared ground-truth; we do not evaluate the model's native geometry consistency on it.

We fit one positive scale per method and scene as the median ratio of reference to native depth over the valid pixels for all five methods. 
At $512\times288$, we back-project the scaled depth $Z_i$ from view $i$ and project the resulting 3D points into view $j$ using the supplied camera intrinsics and poses.
A nearest-depth z-buffer gives the projected camera-axis depth $Z_{i\to j}$. For each shared valid pixel $p$, we compute the symmetric relative depth error
\begin{equation}
    e_{i\to j}(p) =
    \frac{\lvert Z_{i\to j}(p)-Z_j(p)\rvert}
    {\max(Z_{i\to j}(p),Z_j(p))}.
    \label{eq:supp_native_consistency}
\end{equation}
Here, $Z_j$ is the scaled native depth predicted for view $j$. Each direction uses the intersection of pixels with finite, positive reprojected and target depths across all five methods.
L1 and L5 pair views one and five target indices apart, respectively. We average pixel errors within each direction, the two directions within each pair, and retained pairs within each scene. 
Tab.~\ref{tab:supp_native_consistency} shows that WorldBridge has the lowest error at both lags on all three datasets. 
This verifies that our method can generate more coherent 3D worlds.

\noindent\begin{minipage}{\columnwidth}
    \centering
    \footnotesize
    \setlength{\tabcolsep}{1.2pt}
    \captionsetup{type=table}
    \caption{Cross-frame consistency of the native geometry.}
    \label{tab:supp_native_consistency}
    \begin{tabular*}{\columnwidth}{@{\extracolsep{\fill}}lcccccc@{}}
        \toprule
        & \multicolumn{2}{c}{WorldBridge-Long} & \multicolumn{2}{c}{ScanNet v2} & \multicolumn{2}{c}{Oxford Spires} \\
        \cmidrule(lr){2-3}\cmidrule(lr){4-5}\cmidrule(l){6-7}
        Method & L1 $\downarrow$ & L5 $\downarrow$ & L1 $\downarrow$ & L5 $\downarrow$ & L1 $\downarrow$ & L5 $\downarrow$ \\
        \midrule
        3DRAE~\cite{wei20263drae} & \underline{0.0977} & 0.1890 & 0.0557 & 0.1541 & \underline{0.0696} & 0.2412 \\
        Gen3R~\cite{Huang2026Gen3R3S} & 0.1220 & 0.2538 & 0.0962 & 0.2124 & 0.1241 & 0.2042 \\
        GAE~\cite{Lu2026GAELA} & 0.1068 & 0.1813 & \underline{0.0400} & \underline{0.0912} & 0.0748 & \underline{0.1488} \\
        GLD~\cite{jang2026gld} & 0.1143 & \underline{0.1797} & 0.0911 & 0.2788 & 0.0898 & 0.2344 \\
        \midrule
        \textbf{Ours} & \textbf{0.0678} & \textbf{0.1381} & \textbf{0.0294} & \textbf{0.0684} & \textbf{0.0449} & \textbf{0.1314} \\
        \midrule
    \end{tabular*}\end{minipage}\par

\section{Discussion}
\label{sec:discussions}

\subsection{Limitations and Failure Cases}
WorldBridge can generate convincing RGB views coupled with dense point maps, while its 3D geometry is not perfectly fine-grained. 
One possible reason is the bottleneck of view-based 3D representation, rather than the 3D native representation.
For example, in the office and street examples in Fig.~\ref{fig:supp_more_sparse_recon}, the desktops and building fronts establish the overall layout, but chair frames and the edges of walls break up into sparse or uneven points. In the future, it is possible to include the 3D native representations~\cite{hui2022nwd, hui2022neuraltemplate, hu2023clipxplore, hu2024nwd, hu2024cnsedit, hu2026cnseditpp, liu2023exim, du2025hnsr, hu2026pegasus}, e.g., structured 3D latent~\cite{trellis} for better structured or fine-grained geometry generation. Also, we may incorporate agent procedure~\cite{li2026codrawagents, chen2026fact, feng2026funcroom} to further enhance the quality of the results. 

\begin{figure*}[!t]
    \centering
    \includegraphics[width=1.\linewidth]{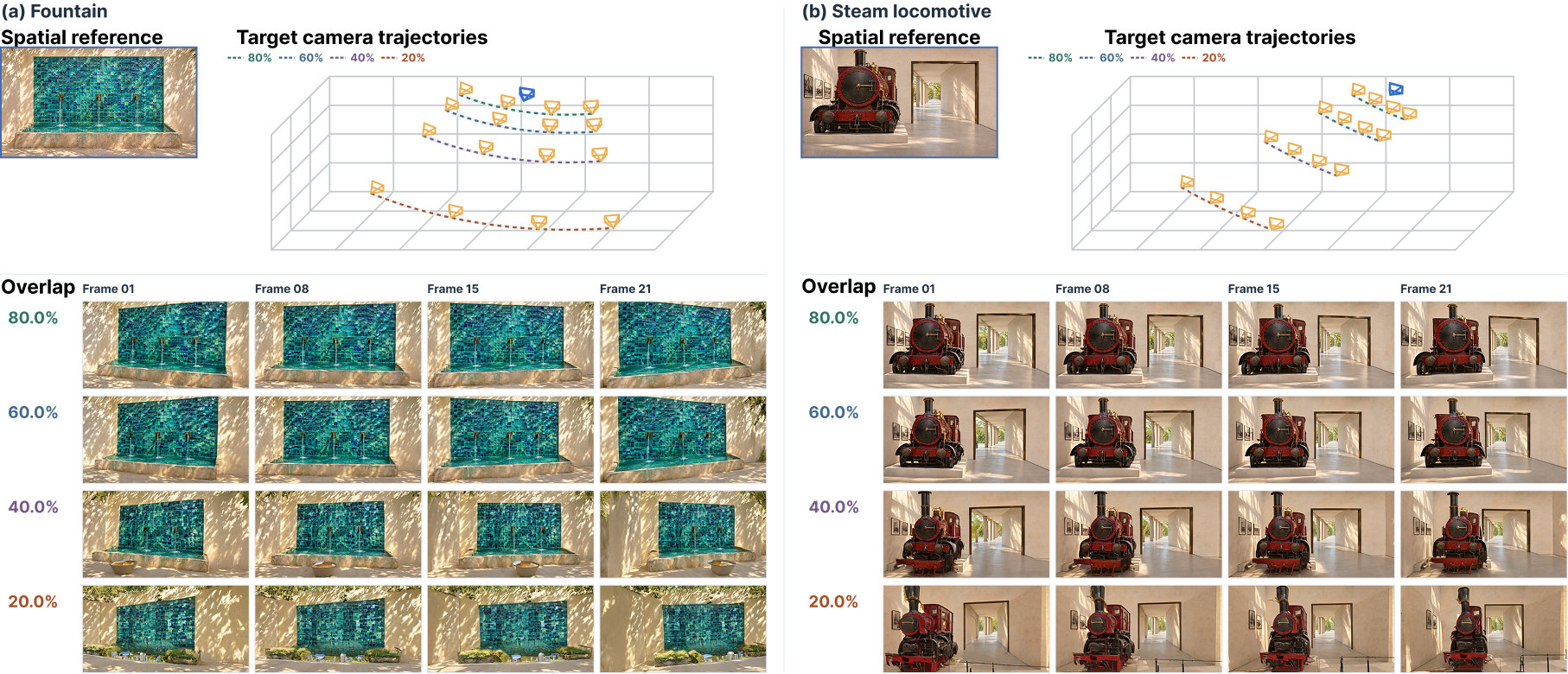}
    \caption{For each spatial reference, four target arcs span 80\%, 60\%, 40\%, and 20\% mean reference-to-target coverage. Columns show the same frame indices along each arc.}
    \label{fig:supp_limitation}
\end{figure*}

A second limitation arises when a target camera sees little of what the spatial reference depicts. 
Fig.~\ref{fig:supp_limitation} keeps one reference fixed per scene and varies the target trajectory. The percentages measure the fraction of the target image visible from the reference camera, averaged along the trajectory.
At 80\% coverage, the fountain surround and locomotive front closely resemble the references. 
At 20\%, the turquoise mosaic remains recognizable, but the surroundings at its base change.
The locomotive is still visible, yet the proportions of its front vary across the sequence. 
As more of the target view falls outside the reference, fewer observed surfaces constrain how the subject fits into its surroundings.
In the future, we can consider more data augmentation and synthesis techniques to provide training data that includes spatial references with low visibility from the target camera trajectories.

\begin{figure*}[!t]
    \centering
    \includegraphics[width=1.\linewidth]{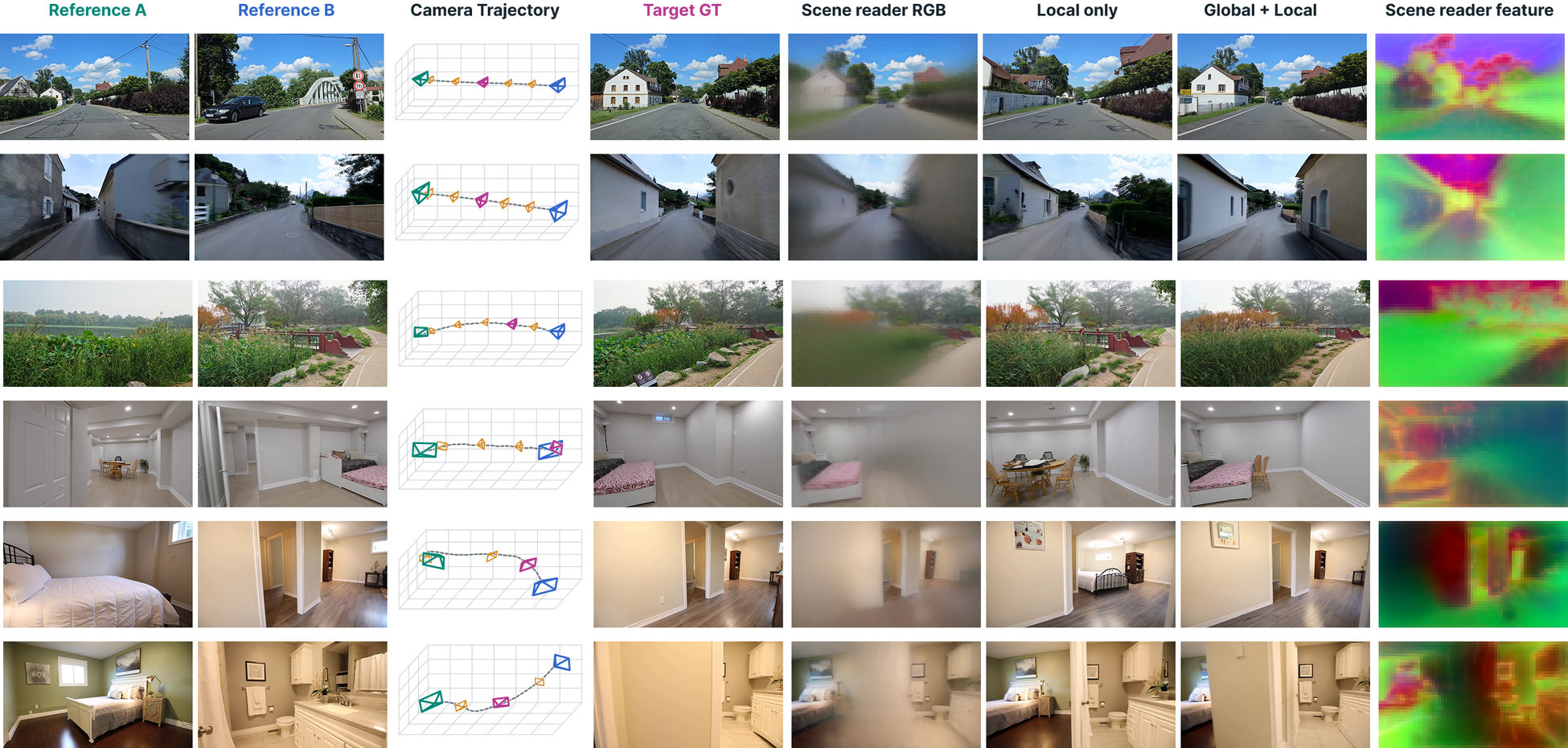}
    \caption{Given two references and a target camera, we compare the held-out target, the scene reader RGB prediction, the local-only output, and the output with both branches. The last column visualizes the scene reader features at the target camera.}
    \label{fig:supp_analysis_global}
\end{figure*}

\subsection{How the Global Branch Helps Generation}
Fig.~\ref{fig:supp_analysis_global} compares the scene reader's RGB prediction with the local-only and full models (local+global) at the same target camera.
The reader queries the shared scene tokens with the target camera rays.
Although its RGB prediction is coarse, it already places the white house to the left of the road in the first row and the bed against the left wall in the fourth.
Broad boundaries in the visualized reader features also follow the road and the walls. These observations suggest that the reader supplies a coarse spatial layout to the requested viewpoint, which provides layout priors for view-specific generation.

The indoor examples show why this layout matters when both references contain plausible but different objects. In the fourth row, the local-only model reproduces the dining set from reference~A at a camera facing the bed in reference~B. In the fifth row, it places a bed beyond a doorway where the target shows a hallway. 
With global context, the bed appears in the first of these views and the hallway stays open in the second.
These comparisons suggest that global conditioning helps determine which observed content belongs at each camera, while the diffusion transformer can use local reference detail to render features such as the patterned bedding.

\twocolumn[{}]
{
    \small
    \bibliographystyle{ieeenat_fullname}
    \bibliography{main_new}
}

\end{document}